\documentclass[article,10pt]{IEEEtran}

\usepackage[utf8]{inputenc} 
\usepackage[T1]{fontenc}
\usepackage{url}              
\usepackage{cite}             

\usepackage[cmex10]{amsmath} 
\usepackage{mleftright}      
\mleftright                  
\usepackage{amsmath}
\DeclareMathOperator*{\argmin}{argmin}
\usepackage{booktabs}         
\usepackage{mathtools}
    \usepackage[absolute]{textpos}
\usepackage{graphicx,cite,amsmath,amssymb,amsfonts,algorithm,mathabx,textcomp,blindtext,xcolor,subfig,float,mwe,dblfloatfix,bm, xfrac,algpseudocode, stackengine, verbatim,xparse, multirow, array, amsthm, arydshln} 

\newtheorem{proposition}{Proposition}

\begin{document}

\begin{textblock}{12}(2,0.5)
\centering
"This work has been submitted to the IEEE for possible publication. Copyright may be transferred without notice, after which this version may no longer be accessible."
\end{textblock}

\title{
Coverage Analysis for Digital Cousin Selection - Improving Multi-Environment Q-Learning
	\thanks{Talha Bozkus and Urbashi Mitra are with the Ming Hsieh Department of Electrical and Computer Engineering, University of Southern California. Email: \{bozkus, ubli\}@usc.edu. Tara Javidi is with the  Department of Electrical and Computer Engineering, University of California at San Diego. 
 Email: \{tjavidi\}@ucsd.edu.}
	\thanks{This work was funded by the following grants:  ARO W911NF1910269, ARO W911NF2410094, NSF CCF-2008927, NSF CCF-2200221, NSF CCF-2311653, NSF A22-2666-S003, ONR N00014-22-1-2363, NSF RINGS-2148313, and is also supported in part by funds from federal agency and industry partners as specified in the Resilient \& Intelligent NextG Systems (RINGS) program).}
}

\author{Talha Bozkus, Tara Javidi, and Urbashi Mitra}
	
\maketitle

\begin{abstract}
Q-learning is widely employed for optimizing various large-dimensional networks with unknown system dynamics. Recent advancements include multi-environment {\bf mixed} $Q$-learning (MEMQ) algorithms, which utilize multiple independent $Q$-learning algorithms across multiple, structurally related but distinct environments and outperform several state-of-the-art $Q$-learning algorithms in terms of accuracy, complexity, and robustness. We herein conduct a comprehensive probabilistic coverage analysis to ensure optimal data coverage conditions for MEMQ algorithms. First, we derive upper and lower bounds on the expectation and variance of different coverage coefficients (CC) for MEMQ algorithms. Leveraging these bounds, we develop a simple way of comparing the utilities of multiple environments in MEMQ algorithms. This approach appears to be near optimal versus our previously proposed partial ordering approach. We also present a novel CC-based MEMQ algorithm to improve the accuracy and complexity of existing MEMQ algorithms. Numerical experiments are conducted using random network graphs with four different graph properties. Our algorithm can reduce the average policy error (APE) by 65\% compared to partial ordering and is 95\% faster than the exhaustive search. It also achieves 60\% less APE than several state-of-the-art reinforcement learning and prior MEMQ algorithms. Additionally, we numerically verify the theoretical results and show their scalability with the action-space size.
\end{abstract}

\begin{IEEEkeywords}
Q-learning, reinforcement learning, coverage coefficient, Markov decision processes (MDPs), random graphs
\end{IEEEkeywords}
\section{Introduction}\label{sec:introduction}

Markov Decision Processes (MDPs) are natural mathematical tools for modeling sequential decision-making problems in various large dimensional networks \cite{barto_sutton_rl, talha_jie_asilomar, mdp_survey}. When the underlying system dynamics, such as transition probabilities and cost functions, are unknown (or non-observable), \textit{model-free} Reinforcement Learning (RL) algorithms such as $Q$-learning can be employed to simulate the system dynamics and learn the policies and value functions for a variety of optimization and network control problems \cite{q_learning_ref, q_learning_ref_2}.

The traditional $Q$-learning algorithm struggles with performance and complexity issues in large MDPs. To address these, various RL algorithms have been proposed. \textit{Ensemble $Q$-learning} algorithms use multiple $Q$-function estimators on a \emph{single} environment for different objectives such as reducing estimation bias and variance \cite{double_q, ensemble_bootstrap_q, maxmin_q}, improving sample efficiency \cite{speedy_q, delayed_q}, enhancing exploration \cite{neural_fitted_q, bootsrapped_dqn}, and handling large state-spaces through approximations \cite{deep_q, q_learning_func_approx}. On the other hand, several algorithms employ \textit{model ensembles}: \cite{modi2020sample} approximates real environments with pre-trained models, \cite{kurutach2018model} improves sample complexity using deep neural network ensembles, \cite{chua2018deep} employs bootstrapped models with probability distributions and \cite{mnih2016asynchronous} employs multiple asynchronous agents that interact with the copies of the environment concurrently. We have recently introduced multi-environment mixed $Q$-learning (MEMQ) algorithms \cite{talha_eusipco, talha_icassp, pn_journal, ln_journal, talha_asilomar}, which demonstrated improved accuracy and complexity performance with respect to the state-of-the-art RL algorithms in large MDPs with a simpler algorithm design. MEMQ algorithms are distinctly characterized by their (i) use of multiple environments and (ii) mixed-learning strategy, which exploits online trajectories as well as synthetically generated trajectories.

The \textbf{first characteristic} of MEMQ algorithms is their use of multiple environments that are synthetically created, distinct but structurally related during learning. These environments are inherently different but share similar characteristics and dynamics, which enable performance improvement with respect to exploration and training by enabling the RL agent to traverse longer trajectories, simulate unobserved trajectories, and identify and leverage patterns in structured environments \cite{pn_journal, ln_journal}. In \cite{talha_asilomar, pn_journal}, $n$-hop environments are utilized, while in \cite{talha_eusipco, talha_icassp, ln_journal}, multiple environments are constructed through the co-link graph transformation \cite{colink_journal}. The use of multiple environments distinguishes MEMQ algorithms from ensemble $Q$-learning algorithms. Furthermore, MEMQ algorithms are tabular $Q$-learning algorithms, avoiding the need for pre-trained models and the complexity of neural networks that pose challenges for model ensemble RL algorithms.

The \textbf{second characteristic} of MEMQ algorithms is their mixed learning strategy -- they combine the features of online and offline RL algorithms and differ from hybrid RL algorithms. In online $Q$-learning, the RL agent learns and updates its $Q$-functions by interacting with the environment in real-time \cite{online_q_learning_1, online_q_learning_2}. In contrast, offline $Q$-learning \cite{offline_q_learning_1, offline_q_learning_2} relies on pre-collected training data. In hybrid RL \cite{hybrid_rl, hybrid_rl_2, hybrid_rl_3, hybrid_rl_4}, the agent has access to an offline dataset and subsequently collects new data by interacting with the environment. On the other hand, in MEMQ algorithms, the original environment is estimated through sampling in real-time (online), and multiple environments are synthetically constructed based on the original environment (offline, but real-time). The synthetic systems run in real-time, in parallel with the original system, with corresponding $Q$-functions for each synthetic system updated at the same time as the real system (online).

To evaluate and compare different RL algorithms effectively, we consider their sample complexities — the number of interactions with the environment needed to learn optimal policies. A lower sample complexity indicates higher efficiency, making RL algorithms practical in scenarios where data collection is expensive or impractical. Sample-efficient \textbf{online RL} algorithms require exploration conditions -- a sufficient number of interactions with the unknown environment is needed, which can be achieved through epsilon-greedy, upper confidence bound or softmax exploration \cite{different_exploration_techniques}. In contrast, sample-efficient \textbf{offline RL} algorithms require data coverage conditions over the offline dataset -- the offline data collection distribution should provide sufficient coverage over the state space \cite{coverage_1, coverage_2}. The work of \cite{coverage_1} shows that coverability conditions -- the existence of a distribution that necessitates uniform coverage of all possible induced state distributions by the data distribution -- can guarantee sample efficiency even in an online setting.

In \textbf{hybrid RL}, combining offline data with online interactions can yield improvements over relying purely on offline or online data \cite{hybrid_rl, hybrid_rl_2, hybrid_rl_3, hybrid_rl_4}. To this end, a new notion of the concentrability coefficient, which quantifies the total coverage of the combined offline and online data, is defined in \cite{hybrid_rl}. In \cite{hybrid_rl_3}, it is shown that having partial concentrability is sufficient to achieve a good sample complexity. Furthermore, \cite{hybrid_rl_4} demonstrates that both computational and statistical efficiency can be achieved if the offline data distribution includes some high-quality policies. MEMQ algorithms combine features of online and offline RL methods; hence, our focus will be on analyzing the coverage conditions of MEMQ algorithms. 

To this end, we will employ a probabilistic framework to derive and analyze the expectation and variance of different coverage coefficients (CCs) of MEMQ algorithms. This is different than the existing work on coverage analysis \cite{coverage_1, coverage_2, coverage_3, coverage_4}, which focuses on deterministic CC analysis. Our probabilistic approach offers several advantages over the deterministic one: (i) there is alignment with the probabilistic $Q$-function analysis in \cite{pn_journal, ln_journal, talha_asilomar}, (ii) the approach ensures a more robust analysis by accounting for noisy or erroneous data collection during sampling and while creating multiple environments, and (iii) we can evaluate different behaviors of CCs under different exploration strategies or structural assumptions on $Q$-functions. We will also present an efficient way of comparing the utility of multiple environments and initializing MEMQ algorithms based on our coverage analysis. While \cite{pn_journal} introduces a heuristic partial ordering approach to compare the utilities of multiple environments based on their $Q$-functions, our novel approach herein is more rigorous and demonstrates higher accuracy and lower complexity than the partial ordering approach and the exhaustive search method. 

We presented our preliminary results in \cite{talha_spawc_paper}, which is a conference paper version of this work. We herein provide significant improvements and novelties with respect to \cite{talha_spawc_paper}: (i) Both lower and upper bounds on different coverage coefficients are presented here whereas \cite{talha_spawc_paper} considers only upper bounds. (ii) The bounds here are generalized to an arbitrary number of actions, whereas \cite{talha_spawc_paper} considers only two actions. (iii) This work focuses on general random networks with different structural properties and analyzes their effect on algorithm performance, while \cite{talha_spawc_paper} focuses on only structured wireless networks. (iv) This work provides an extensive comparison between search methods, prior partial ordering, and the proposed coverage-based ordering in terms of accuracy, complexity, and sensitivity, which is missing in \cite{talha_spawc_paper}. (v) This work considers the effect of different structural assumptions on $Q$-functions and exploration strategies on the performance of the proposed algorithm, which is not considered in \cite{talha_spawc_paper}.

The \textbf{main contributions} of the paper are as follows:

1 -- We adopt a probabilistic approach to derive upper and lower bounds on the expectation and variance of different coverage coefficients (CC) for MEMQ algorithms. 

2 -- We present an efficient way of comparing the utilities of multiple environments in MEMQ algorithms. Unlike prior work on partial ordering \cite{pn_journal}, our approach offers more accurate, low-complexity, and robust initialization for MEMQ algorithms. 

3 -- We present a novel CC-based MEMQ algorithm that modifies the existing MEMQ algorithms to improve their accuracy and complexity. 

4 -- We conduct numerical simulations on random network graphs by changing four different graph properties. Our algorithm can achieve 65\% less average policy error (APE) than the partial ordering technique and 95\% less runtime complexity than the exhaustive search method in finding the optimal control policies. Furthermore, using multiple, synthetic environments suggested by our CC analysis yields 60\% less APE than several state-of-the-art RL and prior MEMQ algorithms. Extensive simulations also validate our assumptions and theoretical analyses.

We use the following notation: the vectors are bold lower case (\textbf{x}), matrices and tensors are bold upper case (\textbf{A}), and sets are in calligraphic font ({$\mathcal{S}$}). Let $\mathbb{E}$ and $\mathbb{V}$ be the expectation and variance operators, respectively.

\section{System Model and Tools}
\label{sec:system_model}

\subsection{Markov Decision Processes (MDPs)}
\label{subsec:mdp}
MDPs are characterized by 4-tuples $\{\mathcal{S}$, $\mathcal{A}$, $p$, $c$\}, where $\mathcal{S}$ and $\mathcal{A}$ denote the finite state and action spaces, respectively. We denote $s_{t}$ as the {\em state} and $a_{t}$ as the {\em action} taken at discrete time period $t$. The probability of transitioning from state $s$ to $s^{\prime}$ under action $a$ is $p_{a}(s,s^{\prime})=p(s^{\prime} = s_{t+1} \mid s = s_{t}, a = a_{t})$, and a bounded average cost $c_{a}(s)= \sum_{s^{\prime} \in \mathcal{S}} p_{a}\left(s, s^{\prime}\right) \hat{c}_{a}(s, s^{\prime})$ is incurred, where $\hat{c}_{a}(s, s^{\prime})$ is the instantaneous transition cost from state $s$ to $s^{\prime}$ under action $a$. We focus on infinite horizon discounted cost MDPs, where $t = \mathbb{Z}^{+}\cup \{0\}$. Our goal is to solve \textit{Bellman's optimality} equation:  
\begin{align}
v^{*}(s)&=\min _{\pi} v_{\pi}(s)=\min _{\pi}\mathbb{E}_{\pi}\left[\sum_{t=0}^{\infty} \gamma^{t} c_{a_{t}}(s_{t}) | s_{0}=s\right],\label{Equ: optimization_eq}\\
\pi^*(s)&=\argmin_{\pi} v_{\pi}(s),\label{Equ: optimization_eq_2}
\end{align}
for all $s \in \mathcal{S}$, where $v_{\pi}$ is the \textit{value function} \cite{barto_sutton_rl} under the \textit{policy} ${\pi}$, $v^{*}$ is the \textit{optimal value function}, $\pi^*$ is the \textit{optimal policy}, and $\gamma \in (0,1) $ is the discount factor. The policy $\pi$ can define either a specific action per state (\textit{deterministic}) or a distribution over the action space per state (\textit{stochastic}) for each time period.

\subsection{$Q$-Learning}\label{subsec:q_learning}

When the system dynamics ($p$ and $c$) are unknown, $Q$-learning can be used to solve (\ref{Equ: optimization_eq}) and (\ref{Equ: optimization_eq_2}). $Q$-learning seeks to find the optimal policy $\pi^{*}$ by learning the $Q$-functions for all $(s,a)$ pairs using the following update rule:
\begin{equation}\label{Equ: $Q$-learning-update-rule}
    Q(s, a) \leftarrow(1-\alpha) Q(s, a)+\alpha(c_{a}(s)+\gamma \min _{a' \in \mathcal{A}} Q(s', a')),
\end{equation}
where $\alpha \in (0,1)$ is the learning rate. In practice, $\epsilon$\textit{-greedy} policies are used to tackle the \textit{exploration-exploitation} trade-off to ensure that sufficient sampling of the system is captured by visiting each state-action pair sufficiently many times \cite{barto_sutton_rl}. To this end, a random action is taken with probability $\epsilon$ (exploration), and a greedy action that minimizes the $Q$-function of the next state is taken with probability $1-\epsilon$ (exploitation). The agent interacts with the environment and collects samples $\{s,a,s',c\}$ to update $Q$-functions using (\ref{Equ: $Q$-learning-update-rule}). The learning strategy must specify the trajectory length ($l$) (the number of states in a trajectory) and the minimum number of visits to each state-action pair ($v$), which is generally used as a termination condition for the sampling operation. $Q$-functions converge to their optimal values with probability one, \textit{\emph{i.e.,}} $Q(s,a) \xrightarrow{\mathit{w.p.1}} Q^{*}(s,a)$ for all $(s,a)$ if necessary conditions are satisfied \cite{q_learning_convergence}. The optimal policy and value functions can be inferred from the $Q$-functions as follows:
\begin{align}
\pi^*(s)=\argmin_{a \in \mathcal{A}} Q^{*}(s, a), \hspace{10pt} v^*(s)=\min_{a \in \mathcal{A}} Q^{*}(s, a).
\end{align}

\subsection{Coverage Coefficient}

The \textit{coverage coefficient} (CC) quantifies how well the offline dataset covers the state space. There exist different definitions for the CC \cite{coverage_2, coverage_3}; we herein use the definition of the CC, which is applicable to discrete and finite state-action spaces and thus can be used with MEMQ algorithms. We denote as follows:
\begin{align}\label{Equ:CC}
    C^{\pi}(s,a) = \frac{d^{\pi}(s,a)}{v(s,a)}.\\
    \quad C^* = \max_{s \in \mathcal{S},a\in\mathcal{A}} C^{\pi}(s,a),
\end{align}
where $C^{\pi}(s,a)$ is the \textit{local CC} for $(s,a)$ under the policy $\pi$, $d^{\pi}(s,a)$ is the occupancy measure under the policy $\pi$ for $(s,a)$, $v(s,a)$ is the exploration distribution for $(s,a)$, and $C^*$ is the \textit{global CC} of the state-action space. When designing or fine-tuning exploration strategies, analyzing the local CC can improve coverage in specific areas of the state-action space, while the global CC is more useful for understanding the overall quality of the dataset when comparing different datasets, algorithms, or exploration strategies. For very large state-action spaces, analyzing the local CC of different state-action samples can provide insights into the global CC, which can be computationally expensive to compute.

Herein, $C^*$ measures the deviation between the exploration distribution $v$ and the distribution induced by the policy $\pi$. If $v$ is close to $d^\pi$, it indicates that the policy $\pi$ spends a comparable amount of time to the exploration of the state-action pair $(s,a)$ by $v$. While achieving perfect coverage ($C^* = 1$) may not be feasible in practice, several techniques may help, including (i) choosing a good exploration distribution $v$, (ii) continuously optimizing the policy $\pi$, (iii) augmenting the original dataset with additional diverse experiences to cover under-explored state-action pairs (which can be achieved through employing multiple environments in MEMQ).

While the distribution $v$ is typically fixed, $d^\pi(s,a)$ changes over time as the policy $\pi$ is optimized. For example, in linear or softmax action-selection strategies, the policy $\pi$ depends on the current $Q$-functions, which change over time \cite{barto_sutton_rl}. Hence, the behavior of $C^\pi(s,a)$ may be non-monotonic over time \emph{i.e.,} it may increase, decrease, or stay constant depending on the relationship between $\pi$ and $v$, which govern the numerator and denominator in (\ref{Equ:CC}), respectively. However, we usually observe a common convergence behavior for different $(s,a)$ pairs as the policy $\pi$ converges.

In the next section, we will theoretically analyze the expectation and variance of different coverage coefficients for MEMQ algorithms.

\section{Theoretical Analysis}

\subsection{Preliminaries and Assumptions}

We use Algorithm 2 of \cite{pn_journal} (n-hop ensemble $Q$-learning = nEQL) as a generic MEMQ algorithm for our analysis; however, our analysis can be generalized to the algorithms of \cite{talha_eusipco, talha_icassp, talha_asilomar, ln_journal}. We first list the assumptions employed throughout the paper:

\textbf{Assumption 1 -- } Let $K$ be the total number of environments in nEQL, and $n$ denote the generic order of the environment. We employ the following distributional assumption on the \textit{$Q$-function errors} of the $n^{th}$ environment:  
\begin{align}
    Q_{t}^{(n)}(s,a) - Q^{*}(s,a) \sim D_n\left(\mu_n, \frac{\lambda_n^2}{3}\right), \label{Equ: distribution_assumption}
\end{align}
where $Q^{*}$ is the optimal $Q$-functions of the original environment ($n=1$), $Q_{t}^{(n)}$ is the $Q$-functions of the $n^{th}$ environment at time $t$, and $D_n$ is some random distribution as a function of $n$. The parameter $\mu_n$ is the \textit{estimation error bias} of $D_n$ and $\frac{\lambda_n^2}{3}$ is the \textit{estimation error variance} of $D_n$. \hfill $\blacksquare$

In prior work \cite{uniform_assump_1, randomized_double_q}, this assumption is used for non-MEMQ algorithms with $D_n$ being uniform or normal and $\mu_n = 0$ for $n=1$ case. In our own work \cite{talha_asilomar, talha_icassp, pn_journal, ln_journal}, this assumption is generalized to $n>1$ with no assumptions on $D_n$ for different MEMQ algorithms.

\textbf{Assumption 2 -- } The distributions $D_n$ are all zero-mean (\emph{i.e.,} $\mu_n = 0$ for all $n$).  \hfill $\blacksquare$

In \cite{pn_journal, ln_journal}, it is shown that $\mu_n$ is close to zero, and $D_n$ can be well-modeled by zero-mean distributions with different but finite variances ($\lambda_n^2 < \infty$) for all $n$. We employ this assumption only to facilitate our theoretical analysis herein; however, it may not always hold true in practice. To this end, the assumption can be relaxed by allowing $\mu_n$ to be non-zero while imposing a constraint that a certain weighted sum of means equals zero, which may be easier to satisfy in practice \cite{pn_journal}.

\begin{figure*}[t]
\small
\begin{align}
    \mathbb{E}\left[\ln C^{\pi^{(n)}}(s,a)\right] \in& \left[
    \ln\frac{1}{1+\theta}-\frac{\lambda^2_n}{6Q^{*}(s,a)^2}-\mathbb{E}\left[\ln v_{sa}\right],\ln\frac{\theta}{1+\theta}+\frac{\lambda^2_n}{3Q^{*}(s,a)^2}
    \left(\frac{2\theta^2}{(1+\theta)^2}-\frac{1}{2}\right)-\mathbb{E}\left[\ln v_{sa}\right]
    \right].\label{Equ: expectation_coverage_n}\\
    \mathbb{V}\left[\ln C^{\pi^{(n)}}(s,a)\right] \in& \left[\max{\{\frac{\lambda_n^2}{3Q^{*}(s,a)^2}(1 - \frac{4\theta}{1+\theta}) + \mathbb{V}\left[\ln v_{sa}\right], 0\}},  \frac{\lambda_n^2}{3Q^{*}(s,a)^2}\left(1+\frac{2\theta}{1+\theta}\right)^2+\mathbb{V}\left[\ln v_{sa}\right]\right].\label{Equ: variance_coverage_n}\\
    \mathbb{E}\left[\ln C^{\pi^{it}}(s,a)\right] \in& \left[
    \ln\frac{1}{1+\theta}-\mathbb{E}\left[\ln v_{sa}\right],\ln\frac{\theta}{1+\theta}+\frac{\lambda^2}{3Q^{*}(s,a)^2}\left(\frac{1-u}{1+u}\right)\left(\frac{2\theta^2}{(1+\theta)^2}\!-\frac{1}{2}\right)-\mathbb{E}\left[\ln v_{sa}\right]
    \right].\label{Equ: expectation_coverage_it}\\
    \mathbb{V}\left[\ln C^{\pi^{it}}(s,a)\right] \in& \left[\mathbb{V}\left[\ln v_{sa}\right],  \frac{\lambda^2}{3Q^{*}(s,a)^2}\left(\frac{1-u}{1+u}\right)\left(1+\frac{2\theta}{1+\theta}\right)^2+\mathbb{V}\left[\ln v_{sa}\right]\right].\label{Equ: variance_coverage_it}\\
    \mathbb{E}\left[\ln C^{\pi^{it}}(s,a)\right] \in& \left[
    \ln\frac{1}{1+\theta}-\mathbb{E}\left[\ln v_{sa}\right],\ln\frac{\theta}{1+\theta}+\frac{1}{Q^{*}(s,a)^2}\frac{f(\lambda,u)}{K}
    \left(\frac{2\theta^2}{(1+\theta)^2}-\frac{1}{2}\right)-\mathbb{E}\left[\ln v_{sa}\right]
    \right].\label{Equ: expectation_coverage_it_K}\\
    \mathbb{V}\left[\ln C^{\pi^{it}}(s,a)\right] \in& \left[\mathbb{V}\left[\ln v_{sa}\right], \frac{1}{Q^{*}(s,a)^2}\frac{f(\lambda,u)}{K}
    \left(1+\frac{2\theta}{1+\theta}\right)^2+\mathbb{V}\left[\ln v_{sa}\right]\right].\label{Equ: variance_coverage_it_K}\\
    \mathbb{E}\left[\ln C^{\pi^{(n)}}(s,a)\right]\in& \left[
    \ln\frac{1}{1\text{+}|\mathcal{A}|\theta-\theta}
- \frac{\lambda_n^2}{6Q^{*}(s,a)^2}
-\mathbb{E}\left[\ln v_{sa}\right],\ln\frac{\theta}{\theta + |\mathcal{A}|-1} \text{+} \frac{\lambda^2_n}{3Q^{*}(s,a)^2}\left(\frac{\frac{|\mathcal{A}|^2}{2}\theta^2}{\left(\theta \text{+} |\mathcal{A}|-1\right)^2} - \frac{1}{2}\right)-\mathbb{E}\left[\ln v_{sa}\right]
\right].\label{Equ: exp_nth_env_atbirary_A}\\
\mathbb{V}\left[\ln C^{\pi^{(n)}}(s,a)\right] \in& \bigg[\max{\{\frac{\lambda_n^2}{3Q^{*}(s,a)^2}(1 - \frac{2|\mathcal{A}|\theta}{1+\theta})+\mathbb{V}\left[\ln v_{sa}\right]},0\},\frac{\lambda_n^2}{3Q^{*}(s,a)^2}\left(1+\frac{|\mathcal{A}|\theta}{\theta + |\mathcal{A}|-1}\right)^2+\mathbb{V}\left[\ln v_{sa}\right]\bigg].\label{Equ: var_nth_env_atbirary_A}
\end{align}
\end{figure*}

\textbf{Assumption 3 -- } We employ a linear action-selection strategy for $d^{\pi}(s,a)$ in (\ref{Equ:CC}):
\begin{align}\label{Equ: linear_action_selection}
    d^{\pi}_t(s,a) = \frac{Q_t(s,a)}{\sum_{i=1}^{|\mathcal{A}|} Q_t(s,a_i)},
\end{align}
where $d^{\pi}_t(s,a)$ is the probability that the policy $\pi$ chooses action $a$ in state $s$ at time $t$, and the $Q$-functions are initialized to very small non-zero values to ensure that (\ref{Equ: linear_action_selection}) is always defined.  \hfill $\blacksquare$

We utilize this assumption to simplify our theoretical analysis. We will later analyze the logarithm of the CC in (\ref{Equ:CC}), and by incorporating (\ref{Equ: linear_action_selection}) into the logarithm of (\ref{Equ:CC}), we will obtain expressions that can be easily computed using (\ref{Equ: distribution_assumption}). We underscore that this assumption is a design choice as one can also employ softmax action-selection, yielding similar theoretical results (see Appendix \ref{Appendix: different_assumption}). On the other hand, the policy $\pi$ was assumed to be deterministic in \cite{pn_journal, ln_journal}, whereas we herein allow $\pi$ to be stochastic.

\textbf{Assumption 4 -- } The optimal $Q$-functions for the same state under different actions have the following relationship:
\begin{align}\label{Equ:structure_of_Q_functions}
    \frac{1}{\theta} \leq \frac{Q^*(s,a_{k_1})}{Q^*(s,a_{k_2})} \leq \theta,
\end{align}
where $\theta \in (1,\infty)$ and $k_1 \neq k_2 \in \mathcal{A}$.   \hfill $\blacksquare$

This assumption allows us to incorporate any prior knowledge of the $Q$-functions through the parameter $\theta$. A small $\theta$ will lead to more stable learning with reduced variance in $Q$-function updates and, consequently, more predictable policy updates and stronger theoretical guarantees. Similar to Assumption 3, this assumption is also a design choice as one can employ different structural assumptions, yielding similar theoretical results (see Appendix \ref{Appendix: different_assumption}).

\textbf{Assumption 5 -- } We have access to a state-action exploration distribution $v$ to sample from that allows us to learn good policies for the first (original) environment.   \hfill $\blacksquare$

This assumption is commonly employed in RL literature \cite{vemula2023virtues, hybrid_rl_2, hybrid_rl_3}. We use the notation $v(s,a) = v_{sa}$.

\textbf{Assumption 6 -- } $D_n$ follows a uniform distribution for all $n$ with (possibly) different $\mu_n$ and $\lambda_n$.  \hfill $\blacksquare$

We employ this assumption only to facilitate our theoretical analysis. However, similar to Assumptions 3 and 4, this assumption is a design choice as the same results can be shown for many other distributions, including triangular, scaled Beta and truncated Normal. \cite{pn_journal}. 

\subsection{Bounds on the coverage coefficients for $|A|=2$}

In this section, we provide key propositions on the expectation and variance of coverage coefficients under different policies under Assumptions 1--5. This analysis helps us understand the convergence behavior of MEMQ algorithms by comparing the simulated CCs with theoretical bounds. Leveraging the theoretical bounds, we can assess the utility of different environments to develop a novel algorithm to improve the accuracy and complexity of the existing MEMQ algorithms. The bounds are given for the logarithm of CC to facilitate analysis due to the ratio of the terms in (\ref{Equ:CC}). We use $C$ for a particular value of CC and $\bar{C}$ for its logarithm.

While we use the basic properties of MEMQ algorithms derived in \cite{pn_journal}, we herein employ different tools and measures to develop our propositions; thus, our analysis is considerably different. We exploit the structure of the linear action-selection strategy and structural assumptions on the Q-functions, leverage Taylor series approximations for random variables, and non-trivial bounds on the ratio of different Q-functions.

\begin{proposition} \label{prop: bounds_on_nth_env}
Let $\pi=\pi^{(n)}$ (the estimated policy of the $n^{th}$ environment in nEQL) in (\ref{Equ:CC}). Then, (\ref{Equ: expectation_coverage_n}) and (\ref{Equ: variance_coverage_n}) hold. (see Appendix \ref{Appendix: proposition_1})
\end{proposition} 
We note that Proposition 1 follows from key definitions of parameters within the algorithm, algebraic manipulations, and classical expectation and variance bounds for positive random variables. We underscore that this proposition considers the individual environments. There are several implications of this proposition. First, by comparing the empirical $\bar{C}$, which changes dynamically over time, with these constant bounds, we can track the convergence of $\bar{C}$. Second, these bounds are governed by the estimation error variance $\lambda^2_n$ (from (\ref{Equ: distribution_assumption})), which varies across environments. As $\lambda^2_n$ decreases, both the expectation and variance bounds become tighter (\emph{i.e.,} the width of the interval decreases). As $\lambda^2_n \rightarrow 0$ and $\theta \rightarrow 1$, the expectation of $\bar{C}$ approaches to $-\ln2 - \mathbb{E}\left[\ln v_{sa}\right]$ and the variance of $\bar{C}$ approaches to $\mathbb{V}\left[\ln v_{sa}\right]$). Third, the parameter $\theta$ is directly related to the smoothness of the underlying cost functions. If the cost of different state-action pairs is similar, the corresponding $Q$-functions also become similar; if the similarity is increased, this leads to a smaller value of $\theta$ and tighter bounds. In contrast, as $\theta \rightarrow \infty$, both upper bounds converge and do not further grow with $\theta$.

We emphasize that these bounds can be further tightened by employing a different action-selection strategy than (\ref{Equ: linear_action_selection}) such as softmax or a different structural assumption on the $Q$-functions than (\ref{Equ:structure_of_Q_functions}) (see Appendix \ref{Appendix: different_assumption} for an example). Furthermore, the quality of the distribution $v$ affects the tightness of the upper and lower bounds equally (\emph{i.e.,} shifts them equally). By adjusting the quality of $v$, we can ensure with high probability that the optimal value of $\bar{C}$ is contained within the interval.

\begin{proposition} \label{prop: bounds_on_it}
Let $\pi=\pi^{it}$ (the estimated ensemble policy of nEQL) in (\ref{Equ:CC}). Then, (\ref{Equ: expectation_coverage_it}) and (\ref{Equ: variance_coverage_it}) hold, where $u \in (0,1)$ is the update ratio of nEQL and $\lambda = \max_n \lambda_n$. (see Appendix \ref{Appendix: proposition_2})
\end{proposition}

The proof of Proposition 2 is a straightforward application of Proposition 1 and properties derived in \cite{pn_journal}. This proposition focuses on the ensemble outcome of nEQL (not individual $Q$-learning algorithms or environments in contrast to Proposition 1). These bounds depend on the hyper-parameter $u$, allowing for optimization through fine-tuning. Following a strategy as in \cite{pn_journal} (choosing a large $u$ or time-dependent $u_t$ such that $u\xrightarrow[]{t \rightarrow \infty}1$) tightens both bounds. For example, as $u \rightarrow 1$ and $\theta \rightarrow 1$, the expectation of $\bar{C}$ approaches to $-\ln2 - \mathbb{E}\left[\ln v_{sa}\right]$ and the variance of $\bar{C}$ approaches to $\mathbb{V}\left[\ln v_{sa}\right]$) as in Proposition 1. The environment with the largest estimation error variance (the largest $\lambda_n$ corresponds to the \textit{worst environment}) governs both bounds; thus, optimizing only the $Q$-learning of the worst environment is an efficient way of improving the overall performance of nEQL. Moreover, even with a large $\lambda$, the adaptive weighting mechanism of nEQL (via update ratio $u$) can tighten the bounds by $u \rightarrow 1$. 

\begin{proposition} \label{prop: bounds_on_it_K}
Let $\pi=\pi^{it}$ (the estimated policy of nEQL) in (\ref{Equ:CC}). Then, (\ref{Equ: expectation_coverage_it_K}) and (\ref{Equ: variance_coverage_it_K}) hold where $f$ is some function of $\lambda$ and $u$.
\end{proposition}

This proposition describes the behavior of $\bar{C}$ as a function of the number of environments in nEQL ($K$). As $K$ increases, both upper bounds tighten, and the variance reduces, which is in line with the variance-reduction property of nEQL (although there is a diminishing return of increasing $K$ with respect to performance improvements \cite{pn_journal}). While Proposition 2 achieves tighter bounds via $u \rightarrow 1$, Proposition 3 accomplishes tighter bounds by increasing $K$. In practice, one can choose either approach depending on whether it is easier to adjust $u$ or $K$. Proving Proposition 3 follows the proof of Proposition 2 herein and properties derived in \cite{pn_journal}.

\subsection{Generalization of the bounds to arbitrary $|A|$}

While the previous bounds hold for $|\mathcal{A}|=2$, they all can be easily generalized to any $|\mathcal{A}|$ using the same approach under Assumptions 1--5. As an example, we generalize Proposition \ref{prop: bounds_on_nth_env} to arbitrary $|\mathcal{A}|$. The proofs for $|A|>2$ are more straightforward to follow, hence their initial presentation for $|A|=2$.
 
\begin{proposition} \label{prop: bounds_on_nth_env_atbirary_A}
Let $\pi=\pi^{(n)}$ (the estimated policy of the $n^{th}$ environment in nEQL) in (\ref{Equ:CC}). Then, the bounds in Proposition \ref{prop: bounds_on_nth_env} can be generalized to arbitrary $|A|$ in (\ref{Equ: exp_nth_env_atbirary_A}) and (\ref{Equ: var_nth_env_atbirary_A}). (see Appendix \ref{Appendix: proposition_4})
\end{proposition}

We underline that these bounds are identical to those in Proposition \ref{prop: bounds_on_nth_env} for $|\mathcal{A}|=2$. For the case $|\mathcal{A}| \gg \theta$, we can show that as $|\mathcal{A}|$ increases, the tightness of both bounds converges and does not further grow with $|\mathcal{A}|$. This result shows that these bounds are scalable with increasing $|\mathcal{A}|$, which is particularly useful as MEMQ algorithms deal with very large state-action spaces. Furthermore, as $\lambda_n^2 \rightarrow 0$ and $\theta \rightarrow 1$, the expectation of $\bar{C}$ approaches to $-\ln|\mathcal{A}|-\mathbb{E}\left[\ln v_{sa}\right]$. For example, if $|\mathcal{A}|=5$ and $v_{sa}=0.2$, the expectation of $C$ converges to 1, which is the optimal value (with some fluctuations around 1 due to the non-zero variance). We leverage the proof of Proposition 1 to prove Proposition 4.

\subsection{Determining the utility of different environments}

The bounds presented in the previous section all rely on the distributional parameters $\lambda_n$ (and $\lambda$). Thus, understanding the relationship between these parameters can be used to improve the accuracy and complexity of nEQL. To this end, our objective is to develop an accurate and computationally efficient way of initializing nEQL. Assume we want to choose $K = 4$ environments to initialize nEQL out of $K_{total}=8$ environments. A challenge is to determine which combination of four environments to include in our algorithm, for example, $\{1^{st},2^{nd},3^{rd},4^{th}\}$ vs $\{1^{st},2^{nd},3^{rd},5^{th}\}$ vs $\{2^{nd},3^{rd},5^{th},6^{th}\}$ and so on. To this end, there are several approaches. Our goal in the current work is to exploit our analysis of the CC to design a new ordering scheme. We shall see that the new ordering based on the CC provides superior performance to our prior approach.
\subsubsection{Exhaustive search}

There are $K_{total} \choose K$ possible ways to select $K$ environments out of $K_{total}$ environments to initialize nEQL. One approach is to evaluate each combination, run nEQL, and choose the set of environments that maximizes policy accuracy. While this method guarantees the most accurate selection, it becomes computationally impractical for large values of $K$ and $K_{total}$.

\subsubsection{Partial ordering}

Proposition 4 of \cite{pn_journal} provides a low-complexity method to compare and select the most informative environments for nEQL by ordering their $Q$-functions. By evaluating the similarity between the $Q$-functions of each environment with that of the original environment, this approach gathers diverse multi-time-scale information and selects the environments that most contribute to the optimal solution.

We illustrate the idea with an example. Let $K_{total} = 8$ and $K=4$. Proposition 4 gives the following partial orderings: $Q^{(1)} \geq Q^{(2)} \geq Q^{(4)} \geq Q^{(8)}$, $Q^{(1)} \geq Q^{(3)} \geq Q^{(6)}$, $Q^{(1)} \geq Q^{(5)}$ and $Q^{(1)} \geq Q^{(7)}$. While these orderings provide comparisons among certain environments, they do not provide a complete comparison of all environments. To address this, we assess the similarity of $Q^{(n)}$ (for $n>1$) to $Q^{(1)}$ based on the given orderings and the idea of exploration-exploitation. Based on given orderings, $Q^{(1)}$ and $Q^{(2)}$ are the most similar, making it a logical choice to include both the $1^{st}$ and $2^{nd}$ environments. The $3^{rd}$ environment is added to explore its potential usefulness, given the uncertain similarity between $Q^{(3)}$ and $Q^{(2)}$. In contrast, $Q^{(4)}$ is less similar to $Q^{(1)}$ than $Q^{(2)}$; thus, we do not include the $4^{th}$ environment, while we can potentially explore more useful environments. Due to the same reason, we also want to explore the $5^{th}$ environment. Hence, a reasonable selection is to use $1^{st},2^{nd},3^{rd}$ and $5^{th}$ environments. While this approach may not always give the most optimal set of environments as opposed to an exhaustive search, it offers a pragmatic and very efficient strategy. 

\subsubsection{Coverage-based ordering}

The exhaustive search method may incur very high computational complexity, whereas the partial ordering approach, being heuristic in nature, may not always yield optimal accuracy. We herein present several results and a novel coverage coefficient-based algorithm to identify the most useful environments for nEQL, which achieves significantly lower complexity than exhaustive search and higher accuracy than partial ordering. The following propositions hold under Assumptions 1--6.

\begin{proposition} \normalfont \label{prop: lambda_1_minimum}
The bounds in Propositions \ref{prop: bounds_on_nth_env} and \ref{prop: bounds_on_nth_env_atbirary_A} are tightest for $n = 1$ and $\lambda_1 = \min_n \lambda_n$. Additionally, $\lambda_n$ is (possibly) non-monotonic across $n$ for $n > 1$. (see Appendix \ref{Appendix: proposition_5})
\end{proposition}

This proposition suggests that the original environment, with the smallest variance (the smallest $\lambda_n$ corresponds to the \textit{best environment}), should always be included in nEQL. By adjusting the parameters of the bounds, we can ensure with a high probability that the optimal value of $\bar{C}$ falls within the interval. This result aligns with the partial ordering idea. On the other hand, the non-monotonicity of $\lambda_n$ for $n>1$ poses challenges in optimizing nEQL, necessitating a strategy to determine the relationship between the $\lambda_n$. We use properties derived in \cite{pn_journal} to prove Proposition 5.

\begin{proposition} \normalfont \label{prop: algorithm_lambda_n}
Let $\lambda_n$ and $\lambda_m$ be the estimation error variance parameters of two distinct environments ($n \neq m$). Let the cost function of the underlying MDP be bounded as $c_a(s) \in [c_{min}, c_{max}]$ for all $s,a$ with $c_{min} > 0$ and $c_{max} < \infty$. Let $\zeta \in (0,1)$ be a weight factor. Then, the following decision rule can be employed to compare $\lambda_n$ and $\lambda_m$:
\begin{align}
&\lambda_n < \lambda_{m} \quad \text{if} \quad f(\gamma, n, m) > \zeta\frac{c_{\text{max}}}{c_{\text{min}}} + (1-\zeta)\frac{c_{\text{min}}}{c_{\text{max}}}. \nonumber\\
&\lambda_n > \lambda_{m} \quad \text{otherwise},
\end{align}
where $f(\gamma, n, m) = \frac{(1-\gamma^{n})(1-\gamma^{m-1})}{(1-\gamma^{m})(1-\gamma^{n-1})}$ and $\gamma \in (0,1)$ is the common discount factor in nEQL. (see Appendix \ref{Appendix: proposition_6})
\end{proposition}

This proposition gives complete ordering between all environments (as opposed to partial ordering). This strategy needs at most ${K_{total}}\choose{2}$ comparisons to order all environments and pick the best $K$ out of them. Clearly, the complexity of choosing the environments is independent of $K$ and significantly reduced compared to the exhaustive search. The function $f$ depends only on the parameter $\gamma$, which makes it robust to the changes in other system parameters. Herein, the minimum and maximum values of the cost function are crucial in the decision rule. If $c_{min}$ is too small or $c_{max}$ is too large, then the rule $\lambda_n > \lambda_m$ may always be chosen, leading to sub-optimal decisions. To this end, one can either design the cost function appropriately or rescale the existing cost values to appropriate values. Herein, the $\zeta$ value is used to handle the cases where the decision rule is inconclusive. Without prior information, $\zeta$ can be simply set to 0.5. However, choosing $\zeta$ randomly in $(0,1)$ or fine-tuning it to minimize the misordering error may yield more accurate results. We leverage the proof of Proposition 5 to prove Proposition 6.

\setlength{\textfloatsep}{5pt}
\algdef{SE}[REPEATN]{RepeatN}{End}[1]{\algorithmicrepeat\ #1 \textbf{times}}{\algorithmicend}
\begin{algorithm}[t]
\caption{Coverage-based ensemble $Q$-learning (CCQ)}
\hspace*{\algorithmicindent} \textbf{Inputs:} MDP model ($\mathcal{S}, \mathcal{A}$, $\hat{p}$, $\hat{c}$), $K$, $K_{total}$, $\gamma$, $u$, $\zeta$ \\
\hspace*{\algorithmicindent} \textbf{Outputs:} $\hat{Q}$, $\hat{\pi}$ 
\begin{enumerate}
\item Find $c_{min}$, $c_{max}$ using the estimated cost functions $\hat{c}$.
\item Sort $\lambda_n$ for $n = 1,2,...,K_{total}$ in increasing order using Proposition \ref{prop: algorithm_lambda_n}.
\item Run nEQL \cite{pn_journal} with the first $K$ environments and $\gamma, u$. Optimize other hyperparameters as in \cite{pn_journal}.
\item Output estimated $Q$-functions $\hat{Q}$ and estimated policy $\hat{\pi}$.
\end{enumerate}
\label{Algo: cc_based_q_learning}
\end{algorithm}

After establishing the bounds on different CCs and a strategy to order the estimation error variances of different environments, we give the main algorithm of this work in Algorithm \ref{Algo: cc_based_q_learning}. The inputs are the MDP model with $\mathcal{S}, \mathcal{A}$ and estimated transition probabilities ($\hat{p}$) and cost functions ($\hat{c}$), the number of environments ($K$) to initialize nEQL out of $K_{total}$ environments, the update ratio of nEQL ($u \in (0,1)$), the discount factor ($\gamma \in (0,1)$) and a weight factor ($\zeta \in (0,1)$). We focus on model-free settings, hence we do not have access to the exact transition probabilities or cost functions. Instead, we collect samples from the environment and estimate transition probabilities and cost functions. To this end, traditional approaches such as sample averaging or data-efficient sampling techniques (see \cite{pn_journal}) for details) can be employed. The parameter $K_{total}$ is chosen moderately large and increases as a function of the state-action space size and $K < K_{total}$. The outputs are the ensemble $Q$-function ($\hat{Q}$) and corresponding estimated policy $\hat{\pi}$. Please refer to \cite{pn_journal} for the details of nEQL.

We emphasize several important points for Algorithm \ref{Algo: cc_based_q_learning}. We initially order the environments based on their estimation error variance in line 2 and prioritize the environments with smaller variances, as this increases the likelihood of picking the environments that achieve the optimal value of the coverage coefficient. As $\lambda_n$ decreases, both expectation and variance bounds tighten. A decrease in the upper bound on the expectation is favorable because an environment with a narrower expectation interval (e.g., [0.36, 1.10]) is more likely to achieve the optimal CC compared to one with a wider interval (e.g., [0.36, 2.72]). Similarly, a decrease in the upper bound on the variance is advantageous because an environment with a smaller variance with the same expectation interval is more likely to achieve the optimal CC than one with a larger variance. We underline that although Algorithm 1 does not explicitly use the bounds on different coverage coefficients, we show that the parameter $\lambda_n$ can be used to determine and sort the utility of multiple environments through these bounds. Thus, they provide a theoretical basis for developing our algorithm.

While the exact bounds in (\ref{Equ: expectation_coverage_n}) -- (\ref{Equ: var_nth_env_atbirary_A}) depend on the optimal $Q$-functions $Q^*$, the value of $\theta$, which depends on $Q^*$, and the distribution $v$, they are \textbf{not needed} in Algorithm \ref{Algo: cc_based_q_learning} for practical purposes. These parameters will \textbf{only} be used to plot the theoretical bounds and compare the simulated CC with the theoretical limits. Similarly, Algorithm \ref{Algo: cc_based_q_learning} does \textbf{not require} the exact values of $\lambda_n^2$;  only the relative orderings are needed, which can be obtained from Proposition \ref{prop: algorithm_lambda_n}.

\section{Numerical Results}
\label{sec:numerical_results}

In this section, we consider a variety of performance metrics to assess the accuracy and complexity performance of Algorithm \ref{Algo: cc_based_q_learning} across different network models.

\subsection{Network models}
We analyze the behavior of CCQ across different random network models by altering four properties of the underlying network graph (\emph{i.e.,} probability transition tensor): structure, sparsity, directionality, and regularity. Each property is changed individually while keeping the others constant. We use the same state-action space size, $|\mathcal{S}| = 10^4$, $|\mathcal{A}| = 4$, and employ a uniform random cost function ($c_{a}(s) \sim \operatorname{unif}(0.5,1)$ for all $s,a$) across all settings. We employ the NetworkX library in Python to create different network graphs.

\subsubsection{Structure}
The first network graph has a block circulant structure, while the second one lacks any structural properties. Both graphs are dense (at least 80\% non-zero entries in the PTT), directed (at least 50\% one-directional connections), and irregular (node degrees vary between 0 and 5).

\subsubsection{Sparsity}
The first network graph is sparse (at least 80\% zero entries in the PTT), while the second one is dense (at least 80\% non-zero entries in the PTT). Both graphs are unstructured, directed, and irregular.

\subsubsection{Directionality}
The first network graph is directed (at least 50\% one-directional connections), while the second one is undirected (all connections are bidirectional). Both graphs are unstructured, dense, and irregular.

\subsubsection{Regularity}

The first network graph is irregular (node degrees vary between 0 and 5), while the second one is regular (all nodes have the same degree). Both graphs are unstructured, dense, and directed.

\subsection{Performance of CCQ across different network models}\label{Subsec:performance_across_different_network_models}

We herein consider three performance metrics to evaluate the performance of CCQ across different network models, as given in the previous section. We first define the \textit{average policy error (APE)} as:

\begin{align}
    APE &=\frac{1}{|\mathcal{S}|} \sum_{s=1}^{|\mathcal{S}|} \mathbf{1}\left(\pi^{*}(s) \neq \hat{\pi}(s)\right),
\end{align}
where $\pi^{*}$ is the optimal policy from (\ref{Equ: optimization_eq_2}), and $\hat{\pi}$ is the estimated policy of CCQ. This metric assesses the accuracy of the estimated policy. Furthermore, we report \textit{computational runtime complexity}, defined as the time it takes to output the estimated policy in Algorithm 1 given the inputs. The complexity measurement of nEQL is detailed along with the average theoretical runtime complexity in \cite{pn_journal}. We also evaluate the \textit{sensitivity} of CCQ across different models by changing the network model parameters in reasonable ranges (for example, 80\% sparsity vs. 70\% sparsity vs. 60\% sparsity or node degree in [0,3] vs. node degree in [0,5], etc.) and measuring the mean percentage change in the APE. We use the following numerical parameters: $|\mathcal{S}|=10^4, |\mathcal{A}|=4, \gamma=0.9$, $v=40$, $l=10$, $K=5$, $K_{total} = 10$, $c_{min} = 0.5$, $c_{max} = 1$, $\zeta = \operatorname{unif}(0,1)$, $\alpha_t = (1+\sfrac{t}{c_1})^{-1}$, $\epsilon_t = \max((c_2)^t, c_3)$, $u_t = 1 - e^{\sfrac{-t}{c_4}}$, where $c_1$, $c_2$, $c_3$, $c_4$ are fine-tuned as in \cite{pn_journal}. Numerical results are shown in Table \ref{table:algorithm_performance}, where each different experiment is run 100 times, and the results are averaged. 

\begin{table}[t]
\centering
\renewcommand{\arraystretch}{1.2} 
\begin{tabular}{|m{1.5cm}|m{1.5cm}|m{1.5cm}|m{1.5cm}|} 
\hline
\textbf{} & \textbf{APE} & \textbf{Runtime (s)} & \textbf{Sensitivity} \\
\hline
\hline
Structured * & $0.06$ & $120$ & $6\%$ \\
\hline
Unstructured & $0.16$ & $340$ & $16\%$ \\
\hline
\hline
Dense & $0.14$ & $300$ & $10\%$ \\
\hline
Sparse * & $0.08$ & $180$ & $6\%$ \\
\hline
\hline
Directed & $0.14$ & $280$ & $8\%$ \\
\hline
Undirected * & $0.10$ & $200$ & $6\%$ \\
\hline
\hline
Regular * & $0.12$ & $180$ & $10\%$ \\
\hline
Irregular & $0.10$ & $200$ & $12\%$ \\
\hline
\end{tabular}
\caption{Performance of CCQ across different network model settings (* denotes the best-performing settings).}
\vspace{-20pt}
\label{table:algorithm_performance}
\end{table}

We observe that the proposed CCQ algorithm performs the best overall for structured, sparse, undirected, and regular networks (denoted by *) by achieving lower APE, lower runtime, and lower sensitivity. In particular, CCQ performs better on structured networks due to their consistent and predictable transition patterns. This predictability facilitates learning and generalization, resulting in a lower APE. Moreover, the structured nature allows for more efficient exploration, leading to faster convergence and reduced runtime. Additionally, the predictability of structured networks ensures less sensitivity to changes in network configurations, resulting in more stable performance. On the other hand, CCQ performs better on sparse networks due to their simplicity, as there are fewer transitions to sample, explore, and learn. This makes learning the $Q$-functions easier and faster. Additionally, sparse networks offer stability, as their limited transitions provide a consistent learning environment that is less affected by fluctuations in network parameters. We emphasize that CCQ shows improvement on undirected and regular networks (versus directed and irregular graphs), yet the impact of directionality and regularity is relatively less significant than the effects of structure and sparsity. Undirected networks offer balanced exploration due to symmetric information flow, while regular networks ensure uniform structures for reliable policy estimations and reduced bias in learning. Overall, the impact of structure is the most significant, followed by sparsity, directionality, and regularity. The best-performing network model is structured, sparse, undirected, regular (S-S-U-R) (achieving 4\% APE with 4\% sensitivity) and the worst-performing one is unstructured, dense, directed, irregular (U-D-D-I) (achieving 18\% APE with 16\% sensitivity).

\subsection{The impact of methodology in choosing the environments}\label{Subsec:performance_across_different_ordering}

In this section, we analyze how the choice of environments impacts the performance of CCQ using three approaches: (i) exhaustive search, (ii) partial ordering \cite{pn_journal}, and (iii) coverage-based ordering (from Proposition 6). To integrate the first two methods into CCQ, we replace step 2 in CCQ with either exhaustive search or partial ordering to select $K$ environments. In order to provide a comprehensive analysis, we conduct simulations on four network configurations: (i) best-performing (S-S-U-R), (ii) structured, dense, directed, regular (S-D-D-R), (iii) unstructured, sparse, undirected, irregular (U-S-U-I), and (iv) worst-performing (U-D-D-I). The numerical parameters 
in Section \ref{Subsec:performance_across_different_network_models} are employed. The numerical results are shown in Table. \ref{table:algorithm_performance_all}.

\begin{table}[t]
\centering
\renewcommand{\arraystretch}{1.2} 
\begin{tabular}{|m{2.1cm}|m{1.5cm}|m{1.6cm}|m{1.5cm}|} 
\hline
\textbf{} & \textbf{APE} & \textbf{Runtime (s)} & \textbf{Sensitivity} \\
\hline
\multicolumn{4}{|c|}{\textbf{Structured-Sparse-Undirected-Regular (S-S-U-R)}} \\
\hline
Exhaustive search & $0.02$ & $1400$ & $0\%$ \\
\hline
Partial ordering & $0.08$ & $80$ & $2\%$ \\
\hline
Coverage-based & $0.04$ & $80$ & $4\%$ \\
\hline
\multicolumn{4}{|c|}{\textbf{Structured-Dense-Directed-Regular (S-D-D-R)}} \\
\hline
Exhaustive search & $0.02$ & $1800$ & $0\%$ \\
\hline
Partial ordering & $0.16$ & $100$ & $4\%$ \\
\hline
Coverage-based & $0.06$ & $140$ & $8\%$ \\
\hline
\multicolumn{4}{|c|}{\textbf{Unstructured-Sparse-Undirected-Irregular (U-S-U-I)}} \\
\hline
Exhaustive search & $0.04$ & $3000$ & $0\%$ \\
\hline
Partial ordering & $0.24$ & $200$ & $4\%$ \\
\hline
Coverage-based & $0.12$ & $240$ & $12\%$ \\
\hline
\multicolumn{4}{|c|}{\textbf{Unstructured-Dense-Directed-Irregular (U-D-D-I)}} \\
\hline
Exhaustive search & $0.06$ & $4800$ & $0\%$ \\
\hline
Partial ordering & $0.30$ & $480$ & $4\%$ \\
\hline
Coverage-based & $0.18$ & $400$ & $18\%$ \\
\hline
\end{tabular}
\caption{Impact of methodology in choosing the environments across different network model settings.}
\label{table:algorithm_performance_all}
\end{table}

While the exhaustive search achieves the smallest APE and has no sensitivity to changes in network models or parameters (as it tries every possible combination of environments), it has a significantly high runtime complexity, which is exacerbated with increasing $K$ and $K_{total}$ (coverage-based is 95\% faster). Hence, it is not preferable for practical purposes. On the other hand, coverage-based ordering can achieve up to 65\% less APE than partial ordering while maintaining similar runtime complexity. Partial ordering exhibits minimal sensitivity as its mechanism is independent of the network model and parameters, but it cannot adaptively change the ordering across different settings, which is a significant drawback. As the underlying network model becomes more complex, coverage-based ordering experiences less increase in APE than partial ordering and less increase in runtime complexity than exhaustive search, making it the most favorable option.

\begin{figure}[t]
    \centering
    \subfloat[Correct detection rates of partial ordering vs coverage-based ordering]{{\includegraphics[width=5cm]{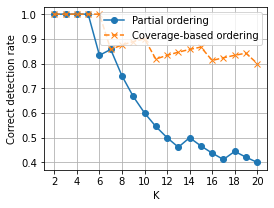}}}%

    \subfloat[CCQ vs other algorithms across increasing state-action space size]{{\includegraphics[width=5cm]{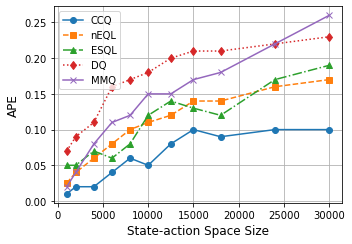}}}%
    \caption{Numerical results.}
    \label{fig:numerical_results}
\end{figure}

We repeat the simulations on the S-S-U-R model with the same parameters, but this time with $K_{total} = 15$ and varying $K$. Table \ref{table:output_environments} shows the order of environments each method outputs as a function of $K$. The exhaustive search always finds the optimal set of environments (at the expense of very high runtime complexity). Coverage-based ordering outputs the optimal set of environments for all $K \leq 6$, with only one mistake for larger $K$, whereas partial ordering makes errors for $K \geq 5$, with errors increasing as $K$ increases. This shows that with a fixed $K_{total}$, the performance of partial ordering declines as $K$ increases, while coverage-based ordering closely follows the exhaustive search and is robust to increase in $K$. This is further validated in Fig. \ref{fig:numerical_results}a, which shows the percentage of optimal environments correctly identified by each method as a function of $K$ ($K \leq 20$) with $K_{total} = 40$. Coverage-based ordering can correctly identify at least 80\% of the optimal environments and appears to be near optimal for larger values of $K$ versus our previously proposed partial ordering approach \cite{pn_journal}, which is one of the major contributions of this work.

\begin{table}[t]
\footnotesize
\centering
\renewcommand{\arraystretch}{1.2} 
\begin{tabular}{|m{0.18cm}|m{2.1cm}|m{2.2cm}|m{2.0cm}|} 
\hline
 K & Exhaustive search & Partial ordering & Coverage-based \\
\hline
\hline
2 & 1,2 & 1,2 & 1,2 \\
\hline
3 & 1,2,3 & 1,2,3 & 1,2,3 \\
\hline
4 & 1,2,3,5 & 1,2,3,5 & 1,2,3,5 \\
\hline
5 & 1,2,3,4,5 & 1,2,3,5,7 & 1,2,3,4,5 \\
\hline
6 & 1,2,3,4,5,7 & 1,2,3,5,7,9 & 1,2,3,4,5,7 \\
\hline
7 & 1,2,3,4,5,7,8 & 1,2,3,5,7,9,11 & 1,2,3,4,5,6,8 \\
\hline
8 & 1,2,3,4,5,7,8,10 & 1,2,3,5,7,9,11,13 & 1,2,3,4,5,6,8,10 \\
\hline
9 & 1,2,3,4,5,7,8,10,14 & 1,2,3,5,7,9,11,13,15 & 1,2,3,4,5,6,8,10,14 \\
\hline
\end{tabular}
\caption{Output environments for different methods on S-S-U-R network model}
\label{table:output_environments}
\end{table}

\subsection{Performance of CCQ across different algorithms}\label{Subsec:performance_across_different_algorithms}

We herein provide a performance comparison between CCQ and similar RL algorithms across increasing state-action space size using the S-S-U-R model. The state-action space is constructed by increasing the state-space size and action-space size comparably (for example, $(|\mathcal{S}|,|\mathcal{A}|)=(1000,2) \rightarrow (2000, 3)$). We employ four algorithms: (i) n-hop Ensemble $Q$-Learning (nEQL) \cite{pn_journal}, (ii) Ensemble-Synthetic-$Q$-Learning (ESQL), (iii) Double $Q$-learning (DQ) and (iv) MaxMin-$Q$-Learning (MMQ). The first two are MEMQ algorithms, while the latter two are ensemble $Q$-learning algorithms. The numerical results in Section \ref{Subsec:performance_across_different_network_models} are employed for CCQ, and the other algorithms are optimized as in \cite{pn_journal, ln_journal}. The APE results are shown in Fig.\ref{fig:numerical_results}b. 

While each algorithm has a comparable APE performance for small state-action spaces, CCQ outperforms other algorithms across larger state-action spaces by achieving 50\% less APE than MEMQ algorithms and 60\% less APE than other ensemble $Q$-learning algorithms. Herein, CCQ performs better than nEQL and ESQL due to its coverage-based environment selection mechanism. In addition, the APE of CCQ converges and does not further grow with the state-action space size, unlike all other algorithms. Similar results can also be observed for the other network model settings.

\begin{figure}[t]
    \tiny
    \centering
    \subfloat[$\ln C^{\pi^{(1)}}(4,0)$ vs bounds \label{Fig:proposition_1_1}]
    {{\includegraphics[width=4.2cm]{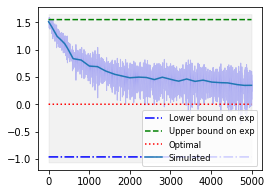}}}\hspace{5pt}
    \subfloat[$\ln C^{\pi^{(2)}}(6,2)$ vs bounds \label{fig:proposition_1_2} ]{{\includegraphics[width=4.05cm]{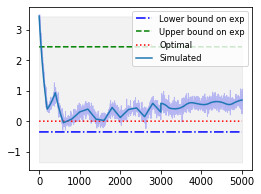}}}

    \subfloat[$\ln C^{\pi^{it}}(4,0)$ (Proposition 2) vs bounds \label{fig:proposition_2} ]{{\includegraphics[width=4.15cm]{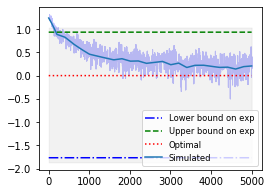}}}\hspace{5pt}
    \subfloat[$\ln C^{\pi^{it}}(4,0)$ (Proposition 3) vs bounds \label{fig:proposition_3} ]{{\includegraphics[width=4.3cm]{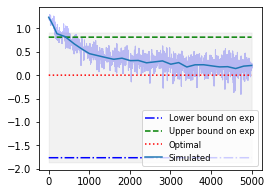}}}
 
    \subfloat[$\ln C^{\pi^{(1)}}(4,0)$ vs upper bounds on expectation across different $|\mathcal{A}|$ \label{fig:upper_bounds_arbitrary_A}]
    {{\includegraphics[width=4.2cm]{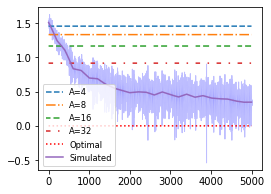}}}  
    \caption{Simulation of theoretical results}
\end{figure}

\subsection{Numerical validation of bounds}

In this section, we compare theoretical results in the propositions with simulation results. We use the numerical parameters in Section \ref{Subsec:performance_across_different_network_models} with $u$ = 0.5. We employ the S-S-U-R network model. In order to plot the bounds, we compute the optimal $Q$-functions $Q^*$ and parameters: $\theta$ = 12, $\lambda_1 = 0.24$, $\lambda_2 = 0.98$, $\lambda_3 = 1.24$, $\lambda_4 = 1.44$, $\lambda_5 = 1.30$ and $\lambda = 1.44$. The parameters $\lambda_n$ are estimated using the sample variance of the $Q$-function errors of individual environments over time. We also numerically verify that Assumption 2 is reasonable as $\mu_1 = 0.02$, $\mu_2 = 0.08$, $\mu_3 = 0.04$, $\mu_4 = 0.08$ and $\mu_5 = 0.10$. The distribution $v$ is obtained by training an independent, robust policy on the original environment by considering different $Q$-function initializations and network parameters (such as $\epsilon_t$ and $\alpha_t$ with different $c_1,c_2,c_3$). We emphasize that this is \textbf{independent} of our learning process.

The simulated $\bar{C}$ of the first and second environments for $(s,a) = (4,0)$ and $(6,2)$, respectively, along with the upper and lower bounds on the expectations from Proposition 1, are shown in Fig.\ref{Fig:proposition_1_1} and Fig.\ref{fig:proposition_1_2}, where the upper and lower bounds of the gray-shaded area are determined by the upper bound on the variance, and the purple-shaded area represents the standard deviation over 100 simulations. We observe that the tightness of the bounds and the behavior of simulated $\bar{C}$ significantly depend on the environment and $(s,a)$ pair. While the upper bound is tighter in the first plot and the lower bound is tighter in the second plot, both expectation and variance bounds hold in the limit. Moreover, $\ln C^{\pi^{(1)}}(4,0)$ smoothly approaches the optimal value 0 (since the distribution $v$ is designed based on this environment) while $\ln C^{\pi^{(2)}}(6,2)$ 
fluctuates and does not converge to 0. The convergence points of simulated $\bar{C}$ for different environments, even for the same $(s,a)$, may potentially differ, but CCQ exploits this fact to achieve faster convergence \cite{pn_journal}. These results focus on two different local CCs, yet similar results can be shown for the other local CCs as well as the global CC.

The simulated $\bar{C}$ under the estimated policy of Algorithm 1 for $(s,a) = (4,0)$, along with the bounds from Proposition 2, are shown in Fig.\ref{fig:proposition_2}. The upper bound on the expectation is significantly tighter than those for individual environments, as CCQ effectively combines exploration across environments. Although the lower bound appears loose, it is less significant since $C$ is always non-negative. The simulated $\bar{C}$ approaches the optimal value faster with less uncertainty (smaller purple-shaded area). This result aligns with the fast convergence of nEQL and Fig.4 of \cite{pn_journal}. The convergence of the simulated $\bar{C}$ to the optimal value also shows that the output policy of Algorithm 1 ($\hat{\pi}$) \textbf{converges} to the optimal policy ($\pi^*$) as this behavior can be observed for all $(s,a)$ pairs.

The same simulated $\bar{C}$, along with the bounds from Proposition 3, are shown in Fig.\ref{fig:proposition_3}. Compared to the bounds in Fig.\ref{fig:proposition_2}, we observe that the upper bounds on both expectation and variance are slightly tighter while the lower bounds are comparable. These bounds implicitly depend on parameters $\lambda$ and $u$ through the function $f$. The bounds are given for $K=5$, and they further tighten as $K$ increases. However, beyond $K \approx 12$ due to the diminishing return, the improvement in the tightness becomes negligible, while APE and runtime complexity increase. On the other hand, Fig.\ref{fig:upper_bounds_arbitrary_A} shows the simulated $\bar{C}$ for $(s,a) = (4,0)$ with $|\mathcal{A}|=2$ and the upper bounds on expectations for different $|\mathcal{A}|$ from Proposition 4. As $|\mathcal{A}|$ doubles, the upper bounds become tighter, which shows their scalability for larger action spaces. While the lower bound slightly loosens, it remains negligible due to the logarithmic term in $|\mathcal{A}|$. We underline that this behavior holds for different $(s,a)$ pairs.

\section{Conclusions}

In this work, we provide a comprehensive coverage analysis of the recent MEMQ algorithms \cite{talha_eusipco, talha_icassp, talha_asilomar, pn_journal, ln_journal} from the probabilistic perspective for optimizing a variety of large random graph networks. We provide upper and lower bounds on the expectation and variance of different coverage coefficients for MEMQ algorithms to ensure a more robust analysis compared to the existing work on coverage analysis. The bounds depend on several network parameters that allow optimization through fine-tuning and tighten with the increasing action-space size. Exploiting these bounds, we present a simple way to order the utility of multiple environments by comparing the estimation error variances of different environments. This approach can identify the optimal set of environments for large values of $K$, outperforming the prior work on partial ordering in terms of accuracy and robustness. Then, we develop a novel coverage-based MEMQ algorithm that improves existing MEMQ algorithms by providing a more accurate and efficient initialization by choosing the optimal set of environments. For numerical experiments, we test our algorithm on random network graphs with four different graph properties. Our algorithm achieves 65\% less APE than the prior partial ordering technique with comparable runtime complexity and is 95\% faster than the exhaustive search. Furthermore, our algorithm achieves 60\% less APE than several RL and prior MEMQ algorithms. We numerically verify that our assumptions and the theoretical bounds on different CCs hold. For future work, we want to theoretically analyze how CCQ performs differently on different network graphs (S-S-U-R vs U-D-D-I) using the singular value decomposition of underlying PTTs. We seek to determine the specific values of $K$ for which different methods yield similar environments. We also work on extending our coverage analysis to multi-agent MEMQ algorithms \cite{talha_icassp25_paper}.

\appendix
\subsection{Proof of Proposition 1}\label{Appendix: proposition_1}

\section{Proof of Proposition 1}

We set $\pi = \pi^{(n)}$. Let $X$ be a RV with mean $\mu > 0$ and variance $\sigma^2 > 0$. We can approximate the distribution of $\ln X$ using the second-order Taylor series expansion as:
\begin{align}
    \mathbb{E}\left[\ln X\right] \approx \ln\mu - \frac{\sigma^2}{2\mu^2},\quad 
    \mathbb{V}\left[\ln X\right] \approx \frac{\sigma^2}{\mu^2}.\label{Equ:exp_var_logX}
\end{align}

These approximations are numerically accurate when $\frac{\mu}{\sigma} > 1.5$. We focus on a generic action $a_k \in \{a_1,a_2\}$ as: $\mathbb{E}\left[\ln d^{\pi^{(n)}}_t(s,a_k)\right] =$
\begin{align}
    =& \mathbb{E}\left[\ln Q_t^{(n)}(s,a_k) - \ln\sum_{i=1}^2Q_t^{(n)}(s,a_i)\right]. \label{Equ:coverage_1}\\
    =& \left(\ln Q^{*}(s,a_k) - \frac{\frac{\lambda_n^2}{3}}{2Q^{*}(s,a_k)^2}\right)- \nonumber\\& \left( \ln \sum_{i=1}^2Q^*(s,a_i)-\frac{\mathbb{V}\left[\sum_{i=1}^2Q_t^{(n)}(s,a_i)\right]}{2\left(\sum_{i=1}^2Q^*(s,a_i)\right)^2}\right),\label{Equ:coverage_3}
\end{align}
where (\ref{Equ:coverage_1}) follows from (\ref{Equ: linear_action_selection}) and the linearity of expectation, and (\ref{Equ:coverage_3}) follows from (\ref{Equ: distribution_assumption}) and (\ref{Equ:exp_var_logX}). Let  $\epsilon_k = \frac{Q^{*}(s,a_k)}{\sum_{i=1}^2Q^*(s,a_i)}$. We first derive the lower bound:
\begin{align}
    \mathbb{E}\left[\ln d^{\pi^{(n)}}_t(s,a_k)\right] \geq& \ln\epsilon_k - \frac{\lambda^2_n}{6Q^{*}(s,a_k)^2}. \label{Equ:coverage_5}\\
    \geq& \ln\frac{1}{1+\theta} - \frac{\lambda^2_n}{6Q^{*}(s,a_k)^2},\label{Equ:coverage_6}
\end{align}
where (\ref{Equ:coverage_5}) follows from algebraic manipulations on (\ref{Equ:coverage_3}), the trivial lower bound on the variance (which is 0), and the definition of $\epsilon_k$, and (\ref{Equ:coverage_6}) follows by the definition of $\epsilon_k$ and (\ref{Equ:structure_of_Q_functions}). We now derive the upper bound:
\begin{align}
    \mathbb{E}[\ln d^{\pi^{(n)}}_t&(s,a_k)] \leq \left(\ln Q^{*}(s,a_k) - \frac{\frac{\lambda_n^2}{3}}{2Q^{*}(s,a_k)^2}\right)- \nonumber\\& \left( \ln \sum_{i=1}^2Q^*(s,a_i)-\frac{\frac{4\lambda_n^2}{3}}{2\left(\sum_{i=1}^2Q_t^*(s,a_i)\right)^2}\right).\label{Equ:coverage_8}\\  
    =&\ln \epsilon_k + \frac{\lambda^2_n}{3Q^{*}(s,a_k)^2}\left(2\epsilon_k^2 - \frac{1}{2}\right). \label{Equ:coverage_10}\\    
    \leq&\ln\frac{\theta}{1+\theta} + \frac{\lambda^2_n}{3Q^{*}(s,a_k)^2}\left(\frac{2\theta^2}{(1+\theta)^2} - \frac{1}{2}\right), \label{Equ:coverage_11}    
\end{align}
where (\ref{Equ:coverage_8}) follows from (\ref{Equ:coverage_3}) and the upper bound on the variance of the sum of dependent RVs (using $\mathbb{V}\left[X+Y\right]\leq\mathbb{V}\left[X\right]+\mathbb{V}\left[Y\right]+2\sqrt{\mathbb{V}\left[X\right]\mathbb{V}\left[Y\right]}$), (\ref{Equ:coverage_10}) follows from the definition of $\epsilon_k$ and (\ref{Equ:coverage_11}) follows from the definition of $\epsilon_k$ and (\ref{Equ:structure_of_Q_functions}). Using (\ref{Equ:CC}), the properties of the expectation, and combining (\ref{Equ:coverage_6}) and (\ref{Equ:coverage_11}) and generalizing to a generic action $a$, (\ref{Equ: expectation_coverage_n}) follows. The bounds are constant with respect to time; thus, we drop the subscript $t$.

We carry out similar operations for the variance. We first derive the upper bound: $\mathbb{V}\left[\ln d^{\pi^{(n)}}_t(s,a_k)\right] = $
\begin{align}
    =&\mathbb{V}\left[\ln Q_t^{(n)}(s,a_k) - \ln \sum_{i=1}^2 Q_t^{(n)}(s,a_i)\right]. \label{Equ:var_n_1}\\
    \leq& \mathbb{V}\left[\ln Q_t^{(n)}(s,a_k)\right] + \mathbb{V}\left[\ln \sum_{i=1}^2 Q_t^{(n)}(s,a_i)\right] \nonumber + \\ &2\sqrt{\mathbb{V}\left[\ln Q_t^{(n)}(s,a_k)\right]\mathbb{V}\left[\ln \sum_{i=1}^2 Q_t^{(n)}(s,a_i)\right]}.\label{Equ:var_n_2}\\
    =&\frac{\frac{\lambda_n^2}{3}}{Q^{*}(s,a_k)^2} + \frac{\mathbb{V}\left[\sum_{i=1}^2 Q_t^{(n)}(s,a_i)\right]}{\left(\sum_{i=1}^2 Q^*(s,a_i)\right)^2} \nonumber \\ &+ 2\sqrt{\frac{\frac{\lambda_n^2}{3}\mathbb{V}\left[\sum_{i=1}^2 Q_t^{(n)}(s,a_i)\right]}{Q^{*}(s,a_k)^2\left(\sum_{i=1}^2 Q^*(s,a_i)\right)^2}}.\label{Equ:var_n_3} \\
    \leq&\frac{\frac{\lambda_n^2}{3}}{Q^{*}(s,a_k)^2} + \frac{\frac{4\lambda_n^2}{3}}{\left(\sum_{i=1}^2 Q^*(s,a_i)\right)^2} \nonumber \\ &+ 2\sqrt{\frac{\frac{4\lambda_n^4}{9}}{Q^{*}(s,a_k)^2\left(\sum_{i=1}^2 Q^*(s,a_i)\right)^2}}.\label{Equ:var_n_3_temp} \\
    =& \frac{\lambda_n^2}{3Q^{*}(s,a_k)^2}\left(1+4\epsilon_k^2+4\epsilon_k\right),\label{Equ:var_n_5}\\
    \leq& \frac{\lambda_n^2}{3Q^{*}(s,a_k)^2}\left(1+\frac{2\theta}{1+\theta}\right)^2,\label{Equ:var_n_6_temp}
\end{align}
where (\ref{Equ:var_n_1}) follows from (\ref{Equ: linear_action_selection}), (\ref{Equ:var_n_2}) follows from the upper bound on the variance of sum of dependent RVs, (\ref{Equ:var_n_3}) follows from (\ref{Equ:exp_var_logX}), (\ref{Equ:var_n_3_temp}) follows from (\ref{Equ:coverage_8}), (\ref{Equ:var_n_5}) follows from the definition of $\epsilon_k$ and algebraic manipulations, and (\ref{Equ:var_n_6_temp}) follows from the definition of $\epsilon_k$ and (\ref{Equ:structure_of_Q_functions}). We now derive the lower bound: $\mathbb{V}\left[\ln d^{\pi^{(n)}}_t(s,a_k)\right] = $
\begin{align}
=&\mathbb{V}\left[\ln Q_t^{(n)}(s,a_k) - \ln \sum_{i=1}^2 Q_t^{(n)}(s,a_i)\right].\label{Equ:var_n_7}\\
\geq& \mathbb{V}\left[\ln Q_t^{(n)}(s,a_k)\right] + \mathbb{V}\left[\ln \sum_{i=1}^2 Q_t^{(n)}(s,a_i)\right] \nonumber \\ & -2\sqrt{\mathbb{V}\left[\ln Q_t^{(n)}(s,a_k)\right]\mathbb{V}\left[\ln \sum_{i=1}^2 Q_t^{(n)}(s,a_i)\right]}. \label{Equ:var_n_8}\\
\geq& \mathbb{V}\left[\ln Q_t^{(n)}(s,a_k)\right] \nonumber \\& -2\sqrt{\mathbb{V}\left[\ln Q_t^{(n)}(s,a_k)\right]\mathbb{V}\left[\ln \sum_{i=1}^2 Q_t^{(n)}(s,a_i)\right]}.\label{Equ:var_n_9}\\
\geq& \frac{\lambda_n^2}{3Q^{*}(s,a_k)^2} - 2\sqrt{\frac{\lambda_n^2}{3Q^{*}(s,a_k)^2}\frac{4\lambda_n^2}{3(\sum_{i=1}^2Q^{*}(s,a_i))^2}}.\label{Equ:var_n_10}\\
=& \frac{\lambda_n^2}{3Q^{*}(s,a_k)^2}(1 - 4\epsilon_k).\label{Equ:var_n_11}\\
\geq& \frac{\lambda_n^2}{3Q^{*}(s,a_k)^2}\left(1 - \frac{4\theta}{1+\theta}\right),\label{Equ:var_n_12}
\end{align}
where (\ref{Equ:var_n_7}) follows from (\ref{Equ:var_n_1}), (\ref{Equ:var_n_8}) follows from the lower bound on the variance of the sum of RVs, (\ref{Equ:var_n_9}) follows from the trivial lower bound on the variance, (\ref{Equ:var_n_10}) follows from (\ref{Equ:exp_var_logX}) and (\ref{Equ:coverage_8}), (\ref{Equ:var_n_11}) follows from the definition of $\epsilon_k$ and algebraic manipulations, and (\ref{Equ:var_n_12}) follows from the upper bound on $\epsilon_k$. 

Using (\ref{Equ:CC}), the properties of the variance operator, the independence between $v$ and $d_t^{\pi^{(n)}}$, combining (\ref{Equ:var_n_6_temp}) and (\ref{Equ:var_n_12}) and generalizing to a generic action $a$, (\ref{Equ: variance_coverage_n}) follows. We take the maximum of the lower bound and 0 to ensure that the variance is always non-negative. We also drop the subscript $t$.

\subsection{Proof of Proposition 2}\label{Appendix: proposition_2}

We set $\pi = \pi^{it}$. In \cite{pn_journal}, it is shown that: 
\begin{align}
    \lim_{t \rightarrow \infty}\mathbb{E}\left[\hat{Q}_t(s,a)\right] = Q^*(s,a). \label{Equ: proof2_1}\\ \lim_{t \rightarrow \infty}\mathbb{V}\left[\hat{Q}_t(s,a)\right] \in \Big[0, \frac{\lambda^2}{3}\frac{1-u}{1+u}\Big].\label{Equ: proof2_2}
\end{align}

Using (\ref{Equ: distribution_assumption}), Assumption 2 and the fact that $Q^*(s,a)$ is deterministic and constant across different $n$: 
\begin{align}
    Q_t^{(n)}(s,a) \sim D_n(Q^*(s,a), \frac{\lambda_n^2}{3}).\label{Equ: proof2_3}
\end{align}

The bounds in (\ref{Equ: expectation_coverage_n}) and (\ref{Equ: variance_coverage_n}) depend on the variance of $Q_t^{(n)}(s,a)$, which is $\frac{\lambda_n^2}{3}$. If we replace $\hat{Q}_t(s,a)$ by $Q_t^{(n)}(s,a)$ in (\ref{Equ: expectation_coverage_n}) and (\ref{Equ: variance_coverage_n}) and let $t \rightarrow \infty$, we can replace the variance $\frac{\lambda_n^2}{3}$ by $\mathbb{V}\left[\hat{Q}_t(s,a)\right]$ using (\ref{Equ: proof2_1}) and (\ref{Equ: proof2_3}). Then, we use (\ref{Equ: proof2_2}) to bound $\mathbb{V}\left[\hat{Q}_t(s,a)\right]$, and the results follow.

\subsection{Proof of Proposition 4}\label{Appendix: proposition_4}

Let $\epsilon_k = \frac{Q^{*}(s,a_k)}{\sum_{i=1}^{|\mathcal{A}|} Q^{*}(s,a_i)}$ similar to Proposition 1. We first derive the lower bound. By following the steps (\ref{Equ:coverage_1}) to (\ref{Equ:coverage_5}), we can show:
\begin{align}
    \mathbb{E}\left[\ln d^{\pi^{(n)}}_t(s,a_k)\right] \geq \ln\epsilon_k - \frac{\lambda_n^2}{6Q^{*}(s,a_k)^2},\label{Equ: proof4_4}
\end{align}

Using the definition of $\epsilon_k$ and (\ref{Equ:structure_of_Q_functions}), we can show that $\epsilon_k \in [\frac{1}{1+|\mathcal{A}|\theta-\theta}, \frac{\theta}{\theta + |\mathcal{A}|-1}]$. Then:
\begin{align}\label{Equ: proof4_5}
    \mathbb{E}\left[\ln d^{\pi^{(n)}}_t(s,a_k)\right] \geq
    \ln\frac{1}{1\!+\!|\mathcal{A}|\theta\!-\!\theta} 
    - \frac{\lambda_n^2}{6Q^{*}(s,a_k)^2}.
\end{align}

We now derive the upper bound: $\mathbb{E}\left[\ln d^{\pi^{(n)}}_t(s,a_k)\right] \leq$
\begin{align}
     &\leq \left(\ln Q^{*}(s,a_k) - \frac{\frac{\lambda_n^2}{3}}{2Q^{*}(s,a_k)^2}\right)- \nonumber\\& \left( \ln \sum_{i=1}^{|\mathcal{A}|}Q^{*}(s,a_i)-\frac{|\mathcal{A}|^2\frac{\lambda_n^2}{3}}{2\left(\sum_{i=1}^{|\mathcal{A}|}Q^{*}(s,a_i)\right)^2}\right).\label{Equ: proof4_6}\\
     &=\ln\frac{Q^{*}(s,a_k)}{\sum_{i=1}^{|\mathcal{A}|}Q^{*}(s,a_i)} + \nonumber \\ &\frac{\lambda^2_n}{3Q^{*}(s,a_k)^2}\left(\frac{\frac{|\mathcal{A}|^2}{2}Q^{*}(s,a_k)^2}{\left(\sum_{i=1}^{|\mathcal{A}|}Q^{*}(s,a_i)\right)^2} - \frac{1}{2}\right).\label{Equ: proof4_7}\\
     &= \ln\epsilon_k + \frac{\lambda^2_n}{3Q^{*}(s,a_k)^2}\left(\frac{|\mathcal{A}|^2}{2}\epsilon_k^2 - \frac{1}{2}\right).\label{Equ: proof4_8}\\
     &\leq\ln\frac{\theta}{\theta + |\mathcal{A}|-1} + \frac{\lambda^2_n}{3Q^{*}(s,a_k)^2}\left(\frac{\frac{|\mathcal{A}|^2}{2}\theta^2}{\left(\theta + |\mathcal{A}|-1\right)^2} - \frac{1}{2}\right),\label{Equ: proof4_9}       
\end{align}
where (\ref{Equ: proof4_6}) follows by generalizing (\ref{Equ:coverage_8}) to $|\mathcal{A}|$ actions, (\ref{Equ: proof4_7}) follows by algebraic manipulations, (\ref{Equ: proof4_8}) follows by the definition of $\epsilon_k$ and (\ref{Equ: proof4_9}) follows from the upper bound on $\epsilon_k$. Using (\ref{Equ:CC}), the properties of the expectation, and combining (\ref{Equ: proof4_5}) and (\ref{Equ: proof4_9}) and generalizing to a generic action $a$, (\ref{Equ: exp_nth_env_atbirary_A}) follows. We drop the subscript $t$ as before.

We now derive the upper bound on the variance. By following steps (\ref{Equ:var_n_1})-(\ref{Equ:var_n_3}), we can show: $\mathbb{V}\left[\ln d^{\pi^{(n)}}_t(s,a_k)\right] \leq $
\begin{align}
    \leq&\frac{\frac{\lambda_n^2}{3}}{Q^{*}(s,a_k)^2} + \frac{\frac{|\mathcal{A}|^2\lambda_n^2}{3}}{\left(\sum_{i=1}^{|\mathcal{A}|} Q^*(s,a_i)\right)^2} \nonumber \\ &+ 2\sqrt{\frac{\frac{|\mathcal{A}|^2\lambda_n^4}{9}}{Q^{*}(s,a_k)^2\left(\sum_{i=1}^{|\mathcal{A}|} Q^*(s,a_i)\right)^2}}. \label{Equ: proof4_10}  \\
    =& \frac{\lambda_n^2}{3Q^{*}(s,a_k)^2}\Bigg(1+|\mathcal{A}|^2\left(\frac{Q^{*}(s,a_k)}{\sum_{i=1}^{|\mathcal{A}|} Q^*(s,a_i)}\right)^2 \nonumber\\& +2|\mathcal{A}|\left(\frac{Q^{*}(s,a_k)}{\sum_{i=1}^{|\mathcal{A}|} Q^*(s,a_i)}\right)\Bigg).\label{Equ: proof4_11}  \\
    =& \frac{\lambda_n^2}{3Q^{*}(s,a_k)^2}\left(1+|\mathcal{A}|^2\epsilon_k^2+2|\mathcal{A}|\epsilon_k\right).\label{Equ: proof4_12} \\    
    \leq& \frac{\lambda_n^2}{3Q^{*}(s,a_k)^2}\left(1+\frac{|\mathcal{A}|\theta}{\theta + |\mathcal{A}|-1}\right)^2,\label{Equ: proof4_14}  
\end{align}
where (\ref{Equ: proof4_10}) follows from (\ref{Equ: proof4_6}), (\ref{Equ: proof4_11})-(\ref{Equ: proof4_12}) follow from the definition of $\epsilon_k$ and algebraic manipulations, and (\ref{Equ: proof4_14}) follows from the upper bound on $\epsilon_k$. We now derive the lower bound. By following the steps (\ref{Equ:var_n_7})-(\ref{Equ:var_n_9}), we can show: $\mathbb{V}\left[\ln d^{\pi^{(n)}}_t(s,a_k)\right] = $
\begin{align}
    \geq& \frac{\lambda_n^2}{3Q^{*}(s,a_k)^2} - 2\sqrt{\frac{\lambda_n^2}{3Q^{*}(s,a_k)^2}\frac{|\mathcal{A}|^2\lambda_n^2}{3(\sum_{i=1}^2Q^{*}(s,a_i))^2}}.\label{Equ: proof4_15} \\
    =& \frac{\lambda_n^2}{3Q^{*}(s,a_k)^2}\left(1 - 2|\mathcal{A}|\epsilon_k\right).\label{Equ: proof4_16}\\
    \geq& \frac{\lambda_n^2}{3Q^{*}(s,a_k)^2}\left(1 - \frac{2|\mathcal{A}|\theta}{1+\theta}\right)\label{Equ: proof4_17},
\end{align}
where (\ref{Equ: proof4_15}) follows from generalizing (\ref{Equ:var_n_10}) to arbitrary $|\mathcal{A}|$, (\ref{Equ: proof4_16}) follows from the definition of $\epsilon_k$ and algebraic manipulations and (\ref{Equ: proof4_17}) follows from the upper bound on $\epsilon_k$. Using (\ref{Equ:CC}), the properties of the expectation, combining (\ref{Equ: proof4_14}) and (\ref{Equ: proof4_17}) and generalizing to a generic action $a$, (\ref{Equ: var_nth_env_atbirary_A}) follows. We drop the subscript $t$ as before.

\subsection{Proof of Proposition 5}\label{Appendix: proposition_5}

Proposition 3 of \cite{pn_journal} can be expressed for a generic state $s$ under the optimal policy $\pi^*$ as:
\begin{align}\label{Equ: proof5_1}
    |Q^{*}(s) - Q^{(n)}_{\pi^*}(s)| < \frac{\gamma}{1-\gamma^n}\frac{1-\gamma^{n-1}}{1-\gamma}|c^{(n)}_{\pi^*}(s)|,
\end{align}
where $Q^{(n)}_{\pi^*}(s)$ is the $Q$-functions and $|c^{(n)}_{\pi^*}(s)|$ is the cost function of the $n^{th}$ environment under $\pi^*$. 
\begin{align}\label{Equ: proof5_2}
     Q^{*}(s) - Q^{(n)}_{\pi^*}(s) \in \Big[-\frac{\gamma}{1-\gamma^n}\frac{1-\gamma^{n-1}}{1-\gamma}c^{(n)}_{\pi^*}(s), \nonumber\\ \frac{\gamma}{1-\gamma^n}\frac{1-\gamma^{n-1}}{1-\gamma}c^{(n)}_{\pi^*}(s)\Big].
\end{align}

Using the uniformity of $D_n$ and the variance of the uniform distribution, we obtain:
\begin{align}
     \mathbb{V}\left[Q^{*}(s) - Q^{(n)}_{\pi^*}(s)\right] &= \left(c^{(n)}_{\pi^*}(s)\right)^2\left(\frac{\gamma}{1-\gamma^n}\frac{1-\gamma^{n-1}}{1-\gamma}\right)^2.\label{Equ: proof5_3}\\
     &=\frac{\lambda^2_n}{3},\label{Equ: proof5_4}
\end{align}
which follows as (\ref{Equ: distribution_assumption}) holds for all $a$. Herein, the term $\Big(\frac{\gamma}{1-\gamma^n}\frac{1-\gamma^{n-1}}{1-\gamma}\Big)^2$ is 0 for $n=1$, and increases monotonically with $n > 1$. On the other hand, under the optimal policy $\pi^*$, $c^{(1)}(s)$ attains its minimum \emph{i.e.,} $c^{(1)}_{\pi^*}(s) \leq c^{(n)}_{\pi^*}(s)$ for any $n$. Hence, $\lambda_n$ is smallest for $n=1$. For $n > 1$, we do not have any information, so $\lambda_n$ does not necessarily behave monotonically across $n$.

\subsection{Proof of Proposition 6}\label{Appendix: proposition_6}

Using (\ref{Equ: proof5_3}) and (\ref{Equ: proof5_4}), we can express the following ratio test:
\begin{align}\label{Equ: proof5_5}
     \frac{c^{(n)}_{\pi^*}(s)}{c^{(m)}_{\pi^*}(s)}  &\underset{\lambda_m}{\overset{\lambda_n}{\gtrless}}\frac{\frac{\gamma}{1-\gamma^{m}}\frac{1-\gamma^{m-1}}{1-\gamma}}{\frac{\gamma}{1-\gamma^n}\frac{1-\gamma^{n-1}}{1-\gamma}}. \nonumber \\ &= \frac{(1-\gamma^{n})(1-\gamma^{m-1})}{(1-\gamma^{m})(1-\gamma^{n-1})}. \nonumber \\ &= f(\gamma, n, m),
\end{align}
where $f$ is a function of $n,m,\gamma$. Let the cost function be bounded as: $c^{(n)}_{\pi^*}(s) \in [c_{min}, c_{max}]$ with $c_{min} > 0$ and $c_{max} < \infty$ for all $n$. The minimum and maximum values of $\frac{c^{(n)}_{\pi^*}(s)}{c^{(m)}_{\pi^*}(s)}$ are $\frac{c_{min}}{c_{max}}$ and $\frac{c_{max}}{c_{min}}$, respectively. Consequently, we have the following decision rule:
\begin{align}
    \lambda_n < \lambda_m \text{\quad iff \quad} f(\gamma, n, m) > \frac{c_{max}}{c_{min}}. \nonumber \\
    \lambda_n > \lambda_m \text{\quad iff \quad} f(\gamma, n, m) < \frac{c_{min}}{c_{max}}.
\end{align}

We emphasize that this decision rule is inconclusive when $f(\gamma, n, m) < \frac{c_{max}}{c_{min}}$ or $f(\gamma, n, m) > \frac{c_{min}}{c_{max}}$. To address this, we can define a boundary by randomly weighting the two thresholds with $\zeta$ and then make a decision based on whether a point in the inconclusive area lies to the right or left of this boundary as follows:
\begin{align}
    \lambda_n < \lambda_{m} \text{\quad iff \quad} f(\gamma, n, m) > \zeta\frac{c_{max}}{c_{min}} + (1-\zeta)\frac{c_{min}}{c_{max}}\\
    \lambda_n > \lambda_{m} \text{\quad iff \quad} f(\gamma, n, m) < \zeta\frac{c_{max}}{c_{min}} + (1-\zeta)\frac{c_{min}}{c_{max}}
\end{align}

\subsection{Different structural assumption and exploration strategy}\label{Appendix: different_assumption}

We use the following softmax action-selection with 2 actions. We focus on a generic action $a_k$:
\begin{equation}\label{Equ: diff_assump_1}
    d^{\pi}_t(s,a_k) = \frac{e^{Q_t(s,a_k)}}{e^{Q_t(s,a_1)}+e^{Q_t(s,a_2)}}.
\end{equation}

We assume the following relationship holds for $k_1 \neq k_2$:
\begin{equation}\label{Equ: diff_assump_2}
    |Q_t(s,a_{k_1}) - Q_t(s,a_{k_2})| \leq \frac{1}{\theta}.
\end{equation}

If we derive the bounds on the expectation and variance of $\ln C^{\pi^{(n)}}(s,a_k)$, the expression for $\epsilon_k$ changes as:
\begin{align}\label{Equ: diff_assump_3}
    \epsilon_k = \frac{e^{Q^*(s,a_k)}}{e^{Q^*(s,a_1)} + e^{Q^*(s,a_2)}}
\end{align}

Using (\ref{Equ: diff_assump_2}) and (\ref{Equ: diff_assump_3}), one can show that $\epsilon_k \leq \frac{1}{1+e^\frac{-1}{\theta}}$. This upper bound is smaller than the previous one ($\epsilon_k \leq \frac{\theta}{1+\theta}$) for $\theta \geq 2$. Consequently, the bounds will be tighter.

\bibliographystyle{unsrt}
\bibliography{references.bib}

\begin{thebibliography}{10}

\bibitem{barto_sutton_rl}
Richard~S Sutton and Andrew~G Barto.
\newblock {\em Reinforcement learning: An introduction}.
\newblock MIT press, 2018.

\bibitem{talha_jie_asilomar}
Jie Wang, Talha Bozkus, Yao Xie, and Urbashi Mitra.
\newblock Reliable adaptive recoding for batched network coding with burst-noise channels.
\newblock In {\em 2023 57th Asilomar Conference on Signals, Systems, and Computers}, pages 220--224. IEEE, 2023.

\bibitem{mdp_survey}
Mohammad~Abu Alsheikh, Dinh~Thai Hoang, Dusit Niyato, Hwee-Pink Tan, and Shaowei Lin.
\newblock Markov decision processes with applications in wireless sensor networks: A survey.
\newblock {\em IEEE Communications Surveys \& Tutorials}, 17(3):1239--1267, 2015.

\bibitem{q_learning_ref}
Beakcheol Jang, Myeonghwi Kim, Gaspard Harerimana, and Jong~Wook Kim.
\newblock Q-learning algorithms: A comprehensive classification and applications.
\newblock {\em IEEE Access}, 7:133653--133667, 2019.

\bibitem{q_learning_ref_2}
Jesse Clifton and Eric Laber.
\newblock Q-learning: Theory and applications.
\newblock {\em Annual Review of Statistics and Its Application}, 7(1):279--301, 2020.

\bibitem{double_q}
Hado Hasselt.
\newblock Double {Q}-learning.
\newblock {\em Advances in neural information processing systems}, 23, 2010.

\bibitem{ensemble_bootstrap_q}
Oren Peer, Chen Tessler, Nadav Merlis, and Ron Meir.
\newblock Ensemble bootstrapping for {Q}-learning.
\newblock In {\em International Conference on Machine Learning}, pages 8454--8463. PMLR, 2021.

\bibitem{maxmin_q}
Qingfeng Lan, Yangchen Pan, Alona Fyshe, and Martha White.
\newblock Maxmin {Q}-learning: Controlling the estimation bias of {Q}-learning.
\newblock {\em CoRR}, abs/2002.06487, 2020.

\bibitem{speedy_q}
Mohammad Ghavamzadeh, Hilbert Kappen, Mohammad Azar, and R{\'e}mi Munos.
\newblock Speedy {Q}-learning.
\newblock {\em Advances in neural information processing systems}, 24, 2011.

\bibitem{delayed_q}
Alexander~L Strehl, Lihong Li, Eric Wiewiora, John Langford, and Michael~L Littman.
\newblock Pac model-free reinforcement learning.
\newblock In {\em Proceedings of the 23rd international conference on Machine learning}, pages 881--888, 2006.

\bibitem{neural_fitted_q}
Martin Riedmiller.
\newblock Neural fitted {Q} iteration--first experiences with a data efficient neural reinforcement learning method.
\newblock In {\em Machine Learning: ECML 2005: 16th European Conference on Machine Learning, Porto, Portugal, October 3-7, 2005. Proceedings 16}, pages 317--328. Springer, 2005.

\bibitem{bootsrapped_dqn}
Ian Osband, Charles Blundell, Alexander Pritzel, and Benjamin Van~Roy.
\newblock Deep exploration via bootstrapped dqn.
\newblock {\em Advances in neural information processing systems}, 29, 2016.

\bibitem{deep_q}
Volodymyr Mnih, Koray Kavukcuoglu, David Silver, Alex Graves, Ioannis Antonoglou, Daan Wierstra, and Martin~A. Riedmiller.
\newblock Playing atari with deep reinforcement learning.
\newblock {\em CoRR}, abs/1312.5602, 2013.

\bibitem{q_learning_func_approx}
Francisco~S Melo and M~Isabel Ribeiro.
\newblock Q-learning with linear function approximation.
\newblock In {\em Learning Theory: 20th Annual Conference on Learning Theory, COLT 2007, San Diego, CA, USA; June 13-15, 2007. Proceedings 20}, pages 308--322. Springer, 2007.

\bibitem{modi2020sample}
Aditya Modi, Nan Jiang, Ambuj Tewari, and Satinder Singh.
\newblock Sample complexity of reinforcement learning using linearly combined model ensembles.
\newblock In {\em International Conference on Artificial Intelligence and Statistics}, pages 2010--2020. PMLR, 2020.

\bibitem{kurutach2018model}
Thanard Kurutach, Ignasi Clavera, Yan Duan, Aviv Tamar, and Pieter Abbeel.
\newblock Model-ensemble trust-region policy optimization.
\newblock {\em arXiv preprint arXiv:1802.10592}, 2018.

\bibitem{chua2018deep}
Kurtland Chua, Roberto Calandra, Rowan McAllister, and Sergey Levine.
\newblock Deep reinforcement learning in a handful of trials using probabilistic dynamics models.
\newblock {\em Advances in neural information processing systems}, 31, 2018.

\bibitem{mnih2016asynchronous}
Volodymyr Mnih, Adria~Puigdomenech Badia, Mehdi Mirza, Alex Graves, Timothy Lillicrap, Tim Harley, David Silver, and Koray Kavukcuoglu.
\newblock Asynchronous methods for deep reinforcement learning.
\newblock In {\em International conference on machine learning}, pages 1928--1937. PMLR, 2016.

\bibitem{talha_eusipco}
Talha Bozkus and Urbashi Mitra.
\newblock Ensemble link learning for large state space multiple access communications.
\newblock In {\em 2022 30th European Signal Processing Conference (EUSIPCO)}, pages 747--751, 2022.

\bibitem{talha_icassp}
Talha Bozkus and Urbashi Mitra.
\newblock Ensemble graph {Q}-learning for large scale networks.
\newblock In {\em ICASSP 2023 - 2023 IEEE International Conference on Acoustics, Speech and Signal Processing (ICASSP)}, pages 1--5, 2023.

\bibitem{pn_journal}
Talha Bozkus and Urbashi Mitra.
\newblock Multi-timescale ensemble $q$-learning for markov decision process policy optimization.
\newblock {\em IEEE Transactions on Signal Processing}, 72:1427--1442, 2024.

\bibitem{ln_journal}
Talha Bozkus and Urbashi Mitra.
\newblock Leveraging digital cousins for ensemble q-learning in large-scale wireless networks.
\newblock {\em IEEE Transactions on Signal Processing}, 72:1114--1129, 2024.

\bibitem{talha_asilomar}
Talha Bozkus and Urbashi Mitra.
\newblock A novel ensemble q-learning algorithm for policy optimization in large-scale networks.
\newblock In {\em 2023 57th Asilomar Conference on Signals, Systems, and Computers}, pages 1381--1386. IEEE, 2023.

\bibitem{colink_journal}
Talha Bozkus and Urbashi Mitra.
\newblock Link analysis for solving multiple-access mdps with large state spaces.
\newblock {\em IEEE Transactions on Signal Processing}, 71:947--962, 2023.

\bibitem{online_q_learning_1}
Chi Jin, Zeyuan Allen-Zhu, Sebastien Bubeck, and Michael~I Jordan.
\newblock Is q-learning provably efficient?
\newblock {\em Advances in neural information processing systems}, 31, 2018.

\bibitem{online_q_learning_2}
Yue Wang and Shaofeng Zou.
\newblock Online robust reinforcement learning with model uncertainty.
\newblock {\em Advances in Neural Information Processing Systems}, 34:7193--7206, 2021.

\bibitem{offline_q_learning_1}
Aviral Kumar, Aurick Zhou, George Tucker, and Sergey Levine.
\newblock Conservative q-learning for offline reinforcement learning.
\newblock {\em Advances in Neural Information Processing Systems}, 33:1179--1191, 2020.

\bibitem{offline_q_learning_2}
Rishabh Agarwal, Dale Schuurmans, and Mohammad Norouzi.
\newblock An optimistic perspective on offline reinforcement learning.
\newblock In {\em International Conference on Machine Learning}, pages 104--114. PMLR, 2020.

\bibitem{hybrid_rl}
Andrew Wagenmaker and Aldo Pacchiano.
\newblock Leveraging offline data in online reinforcement learning.
\newblock In {\em International Conference on Machine Learning}, pages 35300--35338. PMLR, 2023.

\bibitem{hybrid_rl_2}
Ashvin Nair, Murtaza Dalal, Abhishek Gupta, and Sergey Levine.
\newblock Accelerating online reinforcement learning with offline datasets.
\newblock {\em CoRR}, abs/2006.09359, 2020.

\bibitem{hybrid_rl_3}
Tengyang Xie, Nan Jiang, Huan Wang, Caiming Xiong, and Yu~Bai.
\newblock Policy finetuning: Bridging sample-efficient offline and online reinforcement learning.
\newblock {\em Advances in neural information processing systems}, 34:27395--27407, 2021.

\bibitem{hybrid_rl_4}
Yuda Song, Yifei Zhou, Ayush Sekhari, J~Andrew Bagnell, Akshay Krishnamurthy, and Wen Sun.
\newblock Hybrid rl: Using both offline and online data can make rl efficient.
\newblock {\em arXiv preprint arXiv:2210.06718}, 2022.

\bibitem{different_exploration_techniques}
Arryon~D Tijsma, Madalina~M Drugan, and Marco~A Wiering.
\newblock Comparing exploration strategies for q-learning in random stochastic mazes.
\newblock In {\em 2016 IEEE Symposium Series on Computational Intelligence (SSCI)}, pages 1--8. IEEE, 2016.

\bibitem{coverage_1}
Tengyang Xie, Dylan~J Foster, Yu~Bai, Nan Jiang, and Sham~M Kakade.
\newblock The role of coverage in online reinforcement learning.
\newblock {\em arXiv preprint arXiv:2210.04157}, 2022.

\bibitem{coverage_2}
Masatoshi Uehara and Wen Sun.
\newblock Pessimistic model-based offline reinforcement learning under partial coverage.
\newblock {\em arXiv preprint arXiv:2107.06226}, 2021.

\bibitem{coverage_3}
Fanghui Liu, Luca Viano, and Volkan Cevher.
\newblock What can online reinforcement learning with function approximation benefit from general coverage conditions?
\newblock In {\em International Conference on Machine Learning}, pages 22063--22091. PMLR, 2023.

\bibitem{coverage_4}
Haoyi Niu, Yiwen Qiu, Ming Li, Guyue Zhou, Jianming Hu, Xianyuan Zhan, et~al.
\newblock When to trust your simulator: Dynamics-aware hybrid offline-and-online reinforcement learning.
\newblock {\em Advances in Neural Information Processing Systems}, 35:36599--36612, 2022.

\bibitem{talha_spawc_paper}
Talha Bozkus and Urbashi Mitra.
\newblock Coverage analysis of multi-environment q-learning algorithms for wireless network optimization.
\newblock In {\em 2024 IEEE 25th International Workshop on Signal Processing Advances in Wireless Communications (SPAWC)}, pages 376--380, 2024.

\bibitem{q_learning_convergence}
Francisco~S Melo.
\newblock Convergence of {Q}-learning: A simple proof.
\newblock {\em Institute Of Systems and Robotics, Tech. Rep}, pages 1--4, 2001.

\bibitem{uniform_assump_1}
Sebastian Thrun and Anton Schwartz.
\newblock Issues in using function approximation for reinforcement learning.
\newblock In {\em Proceedings of the 1993 Connectionist Models Summer School Hillsdale, NJ. Lawrence Erlbaum}, volume~6, pages 1--9, 1993.

\bibitem{randomized_double_q}
Xinyue Chen, Che Wang, Zijian Zhou, and Keith~W. Ross.
\newblock Randomized ensembled double {Q}-learning: Learning fast without a model.
\newblock {\em CoRR}, abs/2101.05982, 2021.

\bibitem{vemula2023virtues}
Anirudh Vemula, Yuda Song, Aarti Singh, Drew Bagnell, and Sanjiban Choudhury.
\newblock The virtues of laziness in model-based rl: A unified objective and algorithms.
\newblock In {\em International Conference on Machine Learning}, pages 34978--35005. PMLR, 2023.

\bibitem{talha_icassp25_paper}
Talha Bozkus and Urbashi Mitra.
\newblock A multi-agent multi-environment mixed q-learning for partially decentralized wireless network optimization.
\newblock {\em arXiv preprint arXiv:2409.16450}, 2024.

\end{thebibliography}

\end{document}